\documentclass[runningheads]{llncs}

\usepackage{eccv}

\usepackage{eccvabbrv}

\usepackage{graphicx}
\usepackage{booktabs}
\usepackage{multirow}
\usepackage{bm}
\usepackage{algorithm}
\usepackage{algpseudocode}
\newcommand{\noopsort}[1]{}

\usepackage[accsupp]{axessibility}  

\usepackage[pagebackref,breaklinks,colorlinks,citecolor=eccvblue]{hyperref}

\usepackage{orcidlink}

\begin{document}

\title{FMA-Net++: Motion- and Exposure-Aware\\Joint Video Super-Resolution and Deblurring} 
\titlerunning{FMA-Net++}
\author{Geunhyuk Youk\inst{1}\orcidlink{0009-0000-2674-7741} \and
Jihyong Oh\inst{2}\textsuperscript{\textdagger}\orcidlink{0000-0002-1627-0529} \and
Munchurl Kim\inst{1}\textsuperscript{\textdagger}\orcidlink{0000-0003-0146-5419}}
\authorrunning{G.~Youk et al.}
\institute{KAIST, Republic of Korea\\
\email{\{rmsgurkjg, mkimee\}@kaist.ac.kr} \and
CMLab, Chung-Ang University, Republic of Korea\\
\email{jihyongoh@cau.ac.kr}}
\maketitle
\renewcommand{\thefootnote}{}%
\footnotetext{\textsuperscript{\textdagger}~Co-corresponding authors.}
\renewcommand{\thefootnote}{\arabic{footnote}}%
\begin{center}
\vspace{-4mm}
\url{https://kaist-viclab.github.io/fmanetpp\_site/}
\vspace{-6mm}
\end{center}

\begin{figure}[h!]
\centering
    \setlength{\tabcolsep}{0.1cm}
    \begin{tabular}{cc}
        \includegraphics[width=0.58\linewidth]{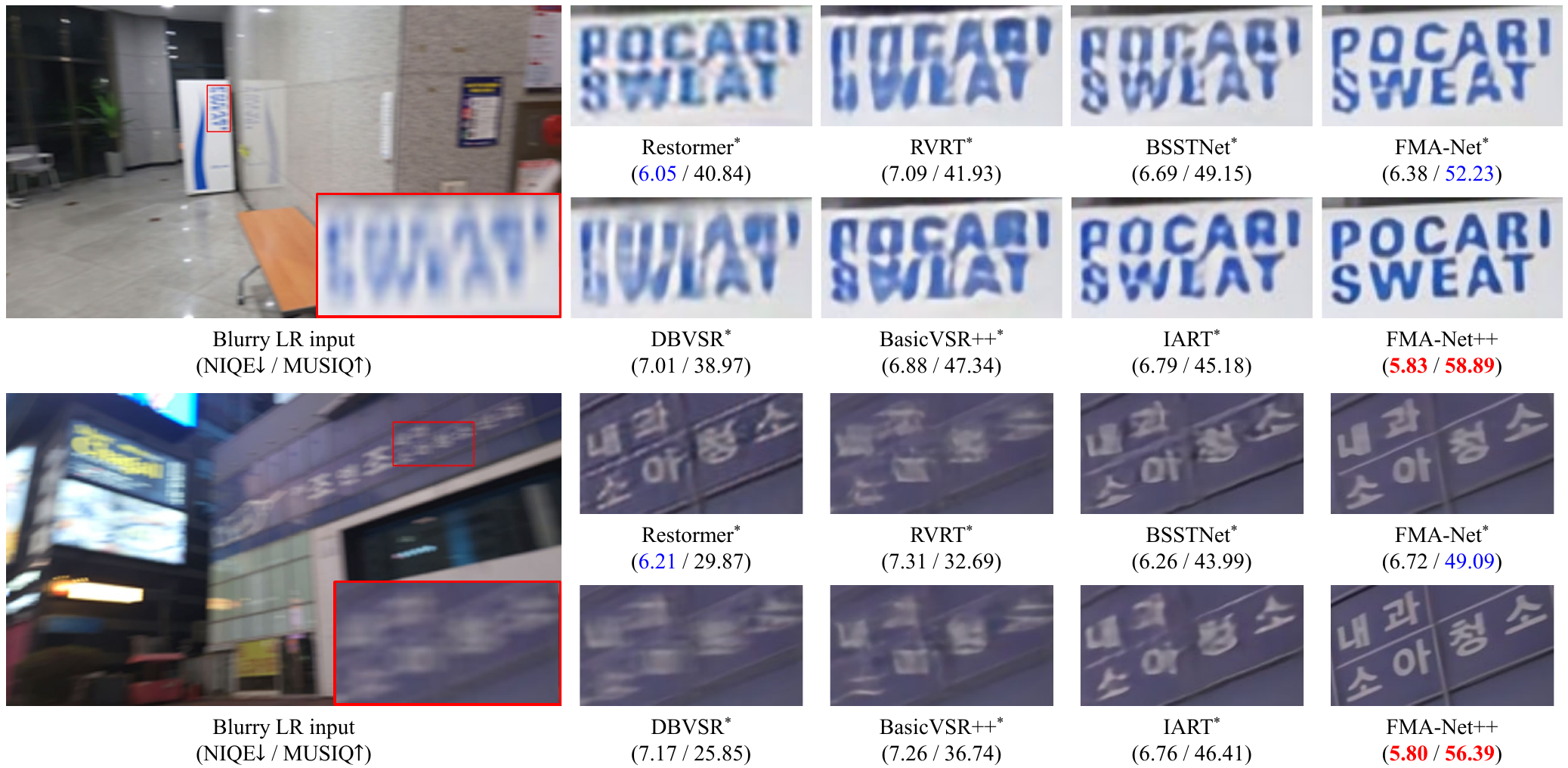} &
        \includegraphics[width=0.35\linewidth]{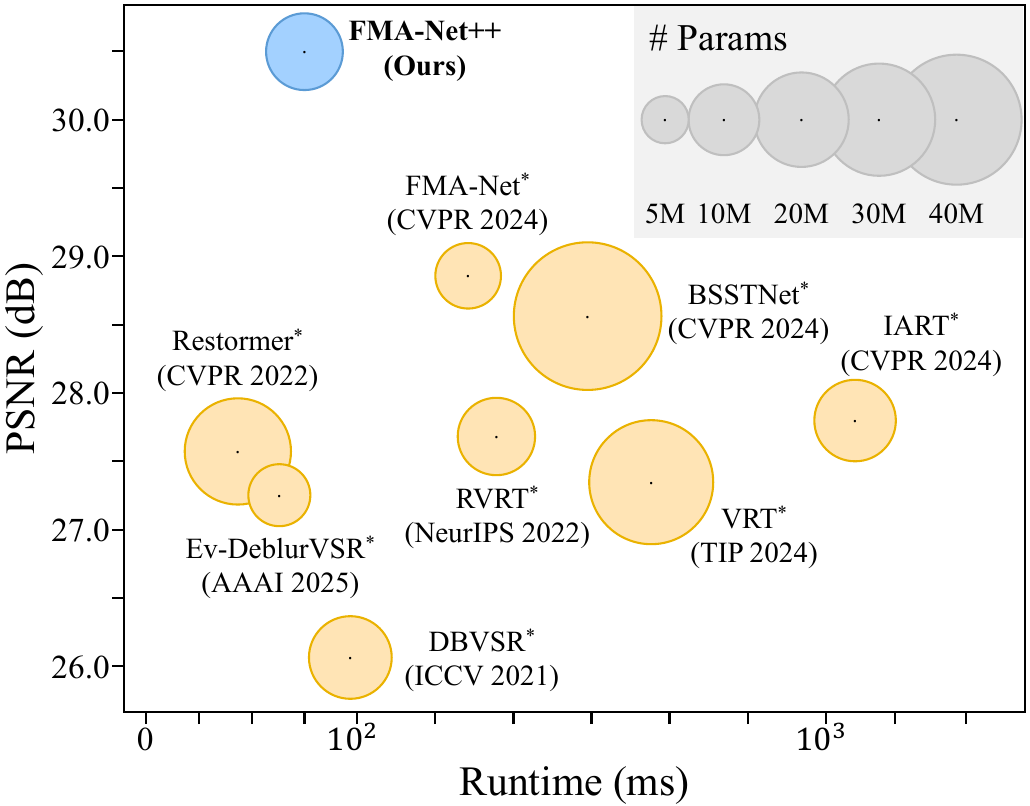} \\
        
        \multicolumn{1}{p{0.58\linewidth}}{\centering \footnotesize (a) Qualitative comparison on challenging real-world videos.} &
        \multicolumn{1}{p{0.4\linewidth}}{\centering \footnotesize (b) Quantitative comparison on the GoPro~\cite{nah2017deep} dataset.}
    \end{tabular}
    \vspace{-2mm}
    \caption{FMA-Net++ outperforms state-of-the-art methods in real-world qualitative results and quantitative benchmarks for joint video super-resolution and deblurring.}
    \label{fig:teaser}
\end{figure}

\vspace{-10mm}
\begin{sloppypar}
\begin{abstract}
Joint video super-resolution and deblurring (VSRDB) requires both efficient long-range temporal modeling and robustness to frame-wise exposure-duration variation, which changes the extent of motion blur across video frames. We propose FMA-Net++, a non-recurrent, sequence-level framework built from Hierarchical Refinement with Bidirectional Aggregation (HRBA) blocks. By stacking HRBA blocks, FMA-Net++ processes video frames in parallel while hierarchically expanding the temporal receptive field, avoiding the limited temporal receptive field of sliding-window designs and the sequential bottleneck of recurrent ones. To handle exposure-duration-dependent blur, we introduce an Exposure Time-aware Modulation (ETM) layer that conditions HRBA features on exposure embeddings from an Exposure Time-aware Feature Extractor (ETE). The conditioned features guide an exposure-aware flow-guided dynamic filtering module to predict motion- and exposure-aware degradation kernels. FMA-Net++ decouples degradation learning from restoration: the former predicts degradation priors and the latter exploits them for efficient high-resolution restoration. To evaluate VSRDB under controlled exposure-duration variation, we introduce the REDS-ME (multi-exposure) and REDS-RE (random-exposure) benchmarks. Trained solely on synthetic data, FMA-Net++ achieves state-of-the-art accuracy and temporal consistency on these benchmarks. It further shows strong out-of-distribution performance on GoPro and challenging real-world videos, while outperforming recent methods in both restoration quality and inference speed.
\end{abstract}
\end{sloppypar}

\section{Introduction} \label{sec:intro}
Joint video super-resolution and deblurring (VSRDB)~\cite{Youk_2024_CVPR,fang2022high,kai2025event} aims to restore sharp high-resolution (HR) videos from blurry low-resolution (LR) inputs. In practice, blurry LR videos are common, and treating SR or deblurring separately is inadequate: SR cannot remove motion blur, while deblurring cannot recover high-frequency details, motivating a joint VSRDB approach~\cite{oh2022demfi, Youk_2024_CVPR}. The physical degradation process underlying these blurry LR videos is driven by two deeply intertwined factors: the \emph{motion field} determines the spatial patterns of blur, and the \emph{exposure time} controls its temporal extent and intensity~\cite{nah2017deep, nah2019ntire, weng2023event}. Compounding this, camera auto-exposure mechanisms vary the exposure dynamically across frames~\cite{kim2022event, weng2023event}. This continuous fluctuation causes the severity of motion blur to change drastically within a single video sequence, resulting in complex, spatio-temporally variant degradations that standard restoration methods struggle to model.

While significant progress has been made in various video restoration tasks~\cite{chan2022basicvsr++, liang2022recurrent, pan2021deep, Youk_2024_CVPR}, most existing methods assume a \textit{fixed exposure time}. This assumption severely limits their robustness, as they struggle to handle the dynamically changing blur severity arising from continuous exposure variations. For instance, VSR~\cite{chan2022basicvsr++, jo2018deep, li2020mucan, chan2021basicvsr, liu2022learning} and video deblurring~\cite{zhang2018adversarial, zhang2022spatio, zhang2024blur, li2023simple, zhong2020efficient} approaches may produce artifacts or temporally inconsistent results when faced with exposure shifts. Even methods designed for unknown degradations, such as Blind VSR~\cite{pan2021deep, bai2024self, lee2021dynavsr}, typically assume spatially-invariant kernels and fail to account for the coupled effect of motion and varying exposure. Furthermore, recent joint VSRDB approaches like FMA-Net~\cite{Youk_2024_CVPR}, despite handling motion-dependent degradation, remain constrained by this fixed-exposure assumption. Thus, VSRDB methods that explicitly address frame-wise exposure-duration variation are critically needed.

\begin{figure}[t]
\centering
\includegraphics[width=\linewidth]{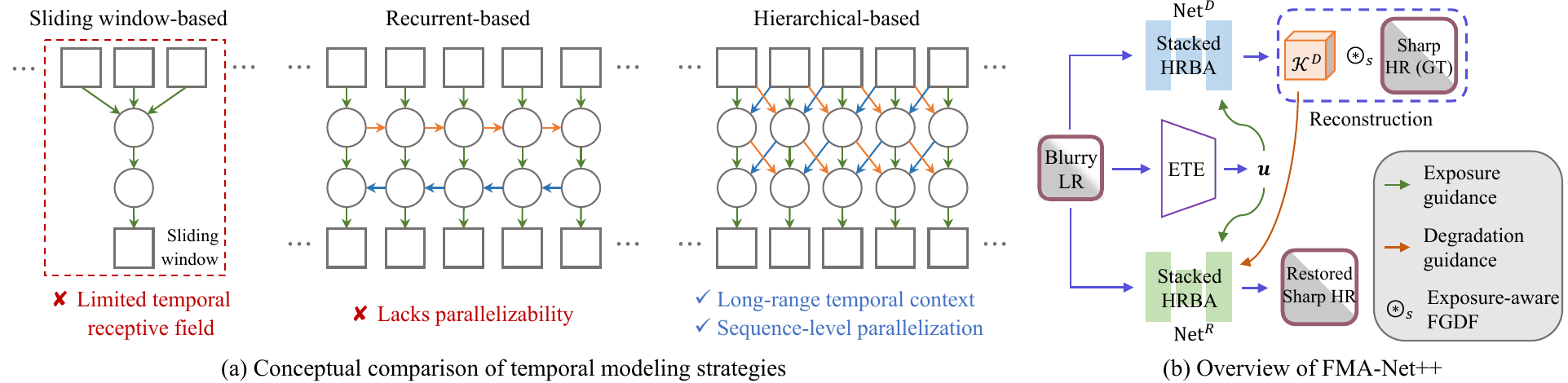}
\vspace{-4mm}
\caption{Conceptual illustration and overview of the FMA-Net++ framework.}
\label{fig:concept}
\vspace{-6mm}
\end{figure}

Moreover, beyond the exposure issue, prevailing temporal modeling strategies face inherent limitations: Sliding-window architectures~\cite{jo2018deep, tian2020tdan, wang2020deep, li2020mucan} tend to suffer from limited temporal receptive fields, while recurrent architectures~\cite{haris2019recurrent, lin2021fdan, chan2021basicvsr, chan2022basicvsr++, liu2022learning} lack parallelizability due to sequential bottlenecks, as conceptually compared in Fig.~\ref{fig:concept}(a). Although recent transformer-based models like VRT~\cite{liang2024vrt} enable parallel processing, they incur heavy computational and memory burdens. To overcome these limits and address the aforementioned exposure variation, we introduce FMA-Net++, an efficient sequence-level framework that couples long-range temporal modeling with exposure-aware degradation estimation.

The core architectural unit of FMA-Net++ is the \textbf{Hierarchical Refinement with Bidirectional Aggregation (HRBA)} block. Instead of relying on restrictive sliding windows, such as those in FMA-Net~\cite{Youk_2024_CVPR}, or inherently sequential recurrent structures~\cite{chan2021basicvsr, liu2022learning, chan2022basicvsr++}, stacking HRBA blocks enables \emph{sequence-level parallelization} and \emph{hierarchically expands the temporal receptive field} to capture long-range dependencies. To handle exposure-duration-dependent blur, each HRBA block includes an \textbf{Exposure Time-aware Modulation (ETM)} layer that conditions features on per-frame exposure embeddings, producing representations rich in both temporal context and exposure information. Leveraging these representations, an exposure-aware Flow-Guided Dynamic Filtering (FGDF) module estimates \emph{motion- and exposure-aware degradation kernels}. Architecturally, we decouple degradation learning from restoration: the former predicts these rich priors, and the latter utilizes them to restore sharp HR frames efficiently, as illustrated in Fig.~\ref{fig:concept}(b).

To systematically evaluate VSRDB under controlled exposure-duration variation, we construct two new benchmarks, REDS-ME (multi-exposure) and REDS-RE (random-exposure). Trained solely on synthetic data, FMA-Net++ achieves state-of-the-art accuracy and temporal consistency on our new benchmarks and the GoPro~\cite{nah2017deep} dataset. It outperforms recent methods in both restoration quality and inference speed, and we further validate robustness on real-world videos using qualitative results and no-reference metrics (see Fig.~\ref{fig:teaser}).

The main contributions of this work are as follows:
\begin{itemize}
    \item We design an efficient, sequence-level architecture based on \textbf{Hierarchical Refinement with Bidirectional Aggregation (HRBA)} blocks. By hierarchically expanding temporal receptive fields without sequential bottlenecks, HRBA effectively captures long-range temporal dependencies while enabling sequence-level parallel processing.
    \item We formulate the physical coupling of motion blur and \emph{frame-wise exposure-duration variation} in VSRDB. To address this, we propose an \textbf{Exposure Time-aware Modulation (ETM)} layer that conditions features on per-frame exposure embeddings, driving the estimation of motion- and exposure-aware degradation kernels.
    \item We introduce two new benchmarks, \textbf{REDS-ME} and \textbf{REDS-RE}, for systematic evaluation under dynamic exposure. Extensive experiments demonstrate that FMA-Net++ achieves state-of-the-art performance, high computational efficiency, and strong generalization to challenging real-world videos.
\end{itemize}

\vspace{-5mm}

\section{Related Work}

\subsection{Joint Video Super-Resolution and Deblurring}
VSRDB tackles the challenging task of jointly restoring sharp HR videos from blurry LR inputs, where blur degradations can be shaped by the coupled effects of motion and exposure. While single-task approaches for VSR~\cite{chan2021basicvsr, chan2022basicvsr++, liang2022recurrent, wang2019edvr, jo2018deep, li2020mucan} or video deblurring~\cite{liang2022recurrent, zhang2018adversarial, zhu2022deep, zhang2024blur, li2023simple} have advanced, applying them sequentially often amplifies artifacts~\cite{oh2022demfi, Youk_2024_CVPR}. However, specific methods tackling this joint VSRDB challenge remain scarce. Early approaches like HOFFR~\cite{fang2022high} struggle with spatially variant blur due to standard CNN limitations. Although FMA-Net~\cite{Youk_2024_CVPR} handles motion-dependent degradation via Flow-Guided Dynamic Filtering, it is constrained by a sliding-window design and a fixed-exposure assumption. FMA-Net++ departs from FMA-Net in two key aspects: it replaces sliding-window restoration with a sequence-level HRBA backbone, and it generalizes degradation estimation to jointly model motion and exposure. Recently, Ev-DeblurVSR~\cite{kai2025event} utilized auxiliary event streams for VSRDB, but it requires non-standard event data and still assumes a fixed exposure time (a limitation explicitly discussed in~\cite{kai2025event}), and therefore does not address RGB-only VSRDB under frame-wise exposure variation. These gaps motivate our sequence-level, exposure-aware approach for robust VSRDB using only standard RGB inputs.

\vspace{-4mm}

\subsection{Temporal Modeling in Video Restoration} \label{sec:temporal_modeling}
Effectively modeling long-range temporal dependencies is crucial for video restoration tasks like VSR. However, prevailing strategies face inherent architectural trade-offs. Sliding-window approaches~\cite{wang2019edvr, jo2018deep, li2020mucan, tao2017detail, Youk_2024_CVPR} operate on fixed local neighborhoods, constraining input flexibility and limiting the capture of long-range context. Conversely, recurrent methods~\cite{chan2021basicvsr, chan2022basicvsr++, liang2022recurrent, liu2022learning} propagate information sequentially. While they enable longer temporal aggregation, they are inherently constrained by sequential processing, which restricts parallelization and makes them prone to vanishing gradients over long sequences~\cite{chiche2022stable, liu2022learning}. Recently, transformer-based models~\cite{cao2021vsrt, liang2024vrt} have been explored. Although VRT~\cite{liang2024vrt} achieves sequence-level parallelization via a shifted-window mechanism~\cite{liu2021swin}, its heavy architecture incurs massive computational and memory costs. Furthermore, most of these works target sharp inputs, lacking robustness to complex degradations. This landscape motivates the need for sequence-level backbones that hierarchically expand temporal receptive fields while enabling efficient parallel processing.

\vspace{-4mm}
\subsection{Exposure Time-Aware Restoration}
In modern camera systems, auto-exposure mechanisms dynamically vary exposure across frames, yielding spatio-temporally variant blur that fixed-exposure models cannot faithfully capture~\cite{kim2022event, weng2023event, shang2023joint}. While recent efforts in related tasks (\eg, video deblurring~\cite{kim2022event, shang2023joint} and frame interpolation~\cite{zhang2020video, weng2023event, shang2023joint}) estimate exposure or exploit auxiliary sensing (events) to guide restoration, they do not explicitly model the \emph{joint} effects of motion and exposure within the VSRDB setting, and event-dependent designs limit practicality for standard RGB videos. To address these limitations, we propose an \textbf{Exposure Time-aware Modulation (ETM)} layer that injects per-frame exposure embeddings into temporal features. These conditioned features then drive our exposure-aware Flow-Guided Dynamic Filtering (FGDF), estimating \emph{motion- and exposure-aware degradation kernels}. This design achieves robust VSRDB using only standard RGB inputs, integrating seamlessly with our sequence-level backbone.

\vspace{-3mm}
\section{Method} \label{sec:method}

\subsection{Problem Formulation} \label{sec:formulation}
We address joint VSRDB under frame-wise varying exposure, where the per-frame exposure time $\Delta t_{e,i}$ is unknown at test time. Given a blurry LR video $\bm{X}=\{\bm{X}_i\}_{i=1}^{T} \in \mathbb{R}^{T \times H \times W \times 3}$, our goal is to restore the sharp HR video $\hat{\bm{Y}}=\{\hat{\bm{Y}}_i\}_{i=1}^{T} \in \mathbb{R}^{T \times sH \times sW \times 3}$ with an upscaling factor $s$.

To motivate an exposure-aware degradation model, we generalize the conventional fixed-exposure blur model~\cite{nah2017deep}. The blurry LR frame $\bm{X}_i$ at a spatial position $\bm{p}$ is physically formed by spatially downsampling and temporally integrating the continuous latent sharp signal $\mathcal{S}$ over the exposure interval $\Delta t_{e,i}$ under the continuous motion field $\bm{M}$:
\begin{equation}
    \bm{X}_i(\bm{p}) = \mathcal{D}_s \left( \frac{1}{\Delta t_{e,i}} \int_{i \cdot \Delta t}^{i \cdot \Delta t + \Delta t_{e,i}} \mathcal{S}\left(\bm{q} + \bm{M}(\bm{q}, \tau), \tau\right) d\tau \right),
\label{eq:physical_degradation}
\end{equation}
where $\mathcal{D}_s$ is the spatial downsampling operator, $\Delta t$ is the frame interval, and $\bm{q}$ is the HR coordinate corresponding to $\bm{p}$. Since directly inverting this continuous physical process is intractable, we approximate it with a discrete, learnable formulation using a spatio-temporally variant degradation kernel $\mathcal{K}_i$:
\begin{equation}
   \bm{X}_i \;\approx\; \mathcal{K}_i *_s \bm{Y}'_i,
   \label{eq:conceptual_degradation}
\end{equation}
where $*_s$ denotes a filtering operation with stride $s$, and $\bm{Y}'_i=\{\bm{Y}_{i-k},\ldots,\bm{Y}_{i+k}\}$ is a short temporal neighborhood of sharp HR frames for a small $k$. Conceptually, the ideal degradation kernel $\mathcal{K}_i$ at $\bm{p}$ depends jointly on the exposure time and the motion field through a complex mapping $\mathcal{F}$:
\begin{equation}
    \mathcal{K}_i(\bm{p}) = \mathcal{F}\left(\Delta t_{e,i}, \left\{\bm{M}(\bm{q},\tau) \mid \bm{q}\in\Omega(\bm{p}); \tau\in[i\cdot\Delta t, i\cdot\Delta t+\Delta t_{e,i}]\right\}\right),
\end{equation}
where $\Omega(\bm{p})$ denotes the spatial neighborhood of HR coordinates corresponding to $\bm{p}$. This formulation captures how dynamic exposure and motion couple to create complex, spatio-temporally variant blur.

Our framework solves this inverse problem of restoring the sharp HR sequence by architecturally decoupling degradation learning from restoration. First, it uses an Exposure Time-aware Feature Extractor and learned optical flow to approximate the properties of $\Delta t_{e,i}$ and $\bm{M}$, thereby estimating a learnable approximation of $\mathcal{K}_i$. Second, these estimated priors guide the restoration network to reconstruct $\hat{\bm{Y}}$.

\begin{figure}[t]
\centering
\includegraphics[width=\linewidth]{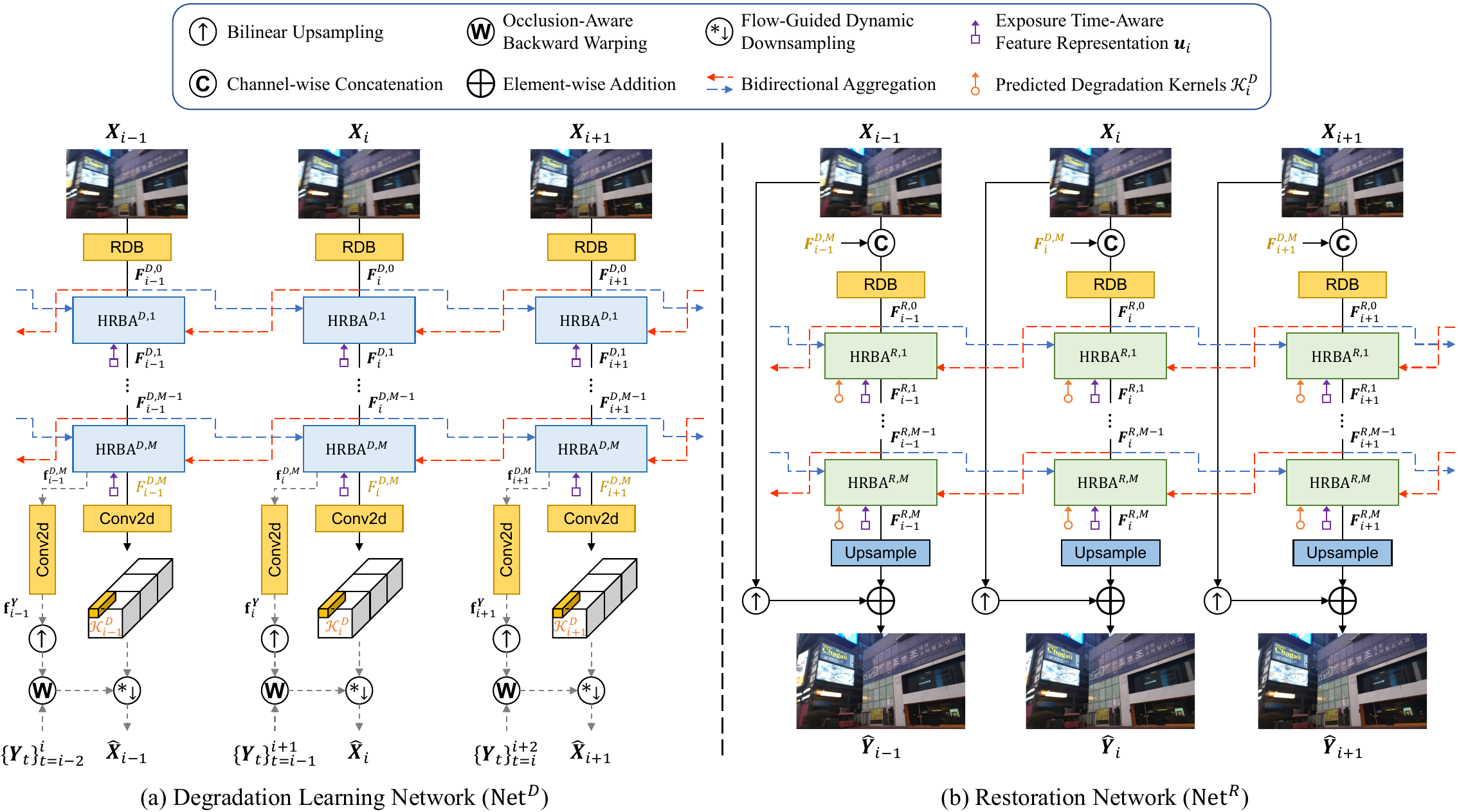}
\caption{Architecture of FMA-Net++ for joint VSRDB.}
\label{fig:network}
\vspace{-5mm}
\end{figure}

\vspace{-4mm}
\subsection{Overall Architecture of FMA-Net++} \label{sec:overview}
Fig.~\ref{fig:network} illustrates the overall architecture of FMA-Net++, which consists of two main networks: a \textbf{Degradation Learning Network} (Net$^D$) and a \textbf{Restoration Network} (Net$^R$). Both networks are built upon stacks of \textbf{Hierarchical Refinement with Bidirectional Aggregation} (HRBA) blocks. This sequence-level backbone processes all frames in parallel, effectively avoiding sequential bottlenecks while hierarchically expanding the temporal receptive field.

Following the degradation formulation in Eq.~\ref{eq:conceptual_degradation}, our architecture decouples degradation learning from restoration. Given an input blurry LR sequence $\bm{X}$ and exposure embeddings from a pretrained Exposure Time-aware Feature Extractor (ETE), Net$^D$ first leverages HRBA and an \textbf{Exposure Time-aware Modulation} (ETM) layer to estimate degradation priors. These priors then drive the exposure-aware Flow-Guided Dynamic Filtering (FGDF) to model spatio-temporally variant degradations. Finally, Net$^R$ restores the sharp HR video $\hat{\bm{Y}}$ guided by these priors, improving both accuracy and efficiency.

\subsection{Hierarchical Refinement with Bidirectional Aggregation} \label{sec:hrba}
As the core architectural unit shared by both Net$^D$ and Net$^R$, the HRBA block overcomes the fundamental trade-offs faced by prior temporal modeling strategies (Sec.~\ref{sec:temporal_modeling}): namely, limited temporal receptive fields in sliding-window methods~\cite{wang2019edvr, Youk_2024_CVPR} and the lack of parallelizability in sequential recurrent approaches~\cite{chan2022basicvsr++, liang2022recurrent}. By stacking HRBA blocks, our architecture enables \emph{sequence-level parallel processing}. At each refinement level, information from increasingly distant past and future frames is aggregated bidirectionally, thus \emph{hierarchically expanding the temporal receptive field} to effectively capture long-range dependencies.

\begin{figure}[t]
\centering
\includegraphics[width=0.5\linewidth]{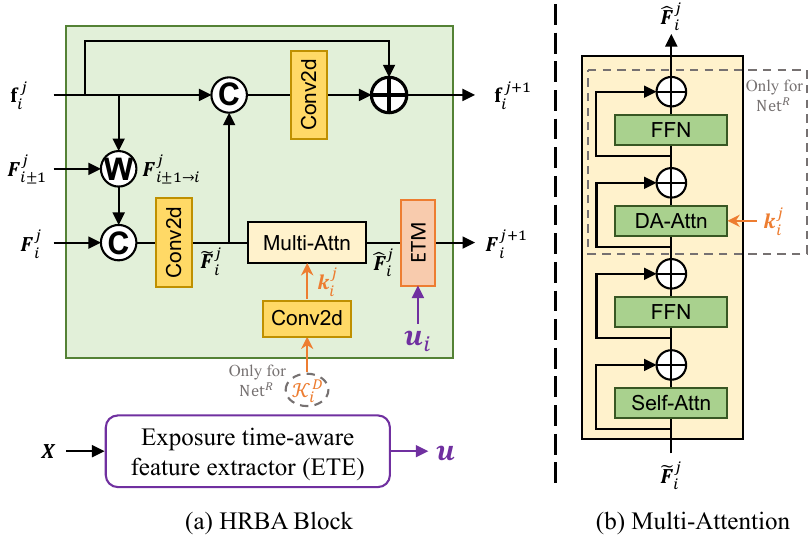}
\vspace{-2mm}
\caption{Details of an HRBA block. (a) Structure of the HRBA block at (\textit{j}+1)-\textit{th} refinement step for \textit{i}-\textit{th} frame (Sec. \ref{sec:hrba}). (b) Structure of Multi-Attention. FFN refers to the feed-forward network of the transformer \cite{vaswani2017attention, dosovitskiy2020image}.}
\label{fig:HRBA}
\vspace{-6mm}
\end{figure}

As shown in Fig.~\ref{fig:HRBA}(a), each HRBA block iteratively refines the feature map $\bm{F}_i^j\in \mathbb{R}^{H \times W \times C}$ and a set of multi-flow-mask pairs $\mathbf{f}_i^j\in \mathbb{R}^{2 \times H \times W \times (2+1)n}$ for a given frame $i$ at refinement step $j+1$. Specifically, $\mathbf{f}_i^j$ is defined as:
\begin{equation}
  \mathbf{f}_i^j \equiv
  \Bigl\{(\bm{f}^{k}_{i \rightarrow i+1},\,\bm{o}^{k}_{i \rightarrow i+1}),
        (\bm{f}^{k}_{i \rightarrow i-1},\,\bm{o}^{k}_{i \rightarrow i-1})\Bigr\}_{k=1}^{n}
\end{equation}
where $n$ is the number of multi-flow-mask pairs, each containing an optical flow $\bm{f}$ and corresponding occlusion mask $\bm{o}$ representing motion towards neighbors $i\pm1$. Keeping multiple motion hypotheses ($n>1$) enhances robustness under severe blur by providing one-to-many correspondences~\cite{chan2021understanding, hu2022many}. The refinement process first computes intermediate features $\Tilde{\bm{F}}_i^j$ via occlusion-aware warping~\cite{jaderberg2015spatial, oh2022demfi} of neighboring features $\bm{F}_{i \pm 1}^j$ using $\mathbf{f}_i^j$, followed by fusion using concatenation and convolution. The multi-flow-mask pairs are then updated residually, $\mathbf{f}_i^{j+1} = \mathbf{f}_i^j + \Delta\mathbf{f}_i^j$, where the residual $\Delta\mathbf{f}_i^j$ is predicted based on $\Tilde{\bm{F}}_i^j$ and $\mathbf{f}_i^j$. The intermediate feature $\Tilde{\bm{F}}_i^j$ is further enhanced through two crucial modules before producing the final output $\bm{F}_i^{j+1}$.

\noindent\textbf{Multi-Attention.} As shown in Fig.~\ref{fig:HRBA}(b), the multi-attention module employs self-attention~\cite{vaswani2017attention} to capture spatial dependencies and integrate the aggregated hierarchical temporal context. Within Net$^R$, it subsequently applies Degradation-Aware (DA) attention. This cross-attention mechanism uses query $\bm{Q}$ derived from the estimated exposure- and motion-aware degradation kernel $\mathcal{K}^{D}_i$ (predicted by Net$^D$), while key $\bm{K}$ and value $\bm{V}$ are projected from the self-attention output. This allows Net$^R$ features to adapt specifically to the estimated degradation characteristics of each frame.

\noindent\textbf{Exposure Time-aware Modulation (ETM).} To handle frame-wise exposure variation, every HRBA block applies ETM via a lightweight SFT layer~\cite{wang2018recovering}. The exposure embedding $\bm{u}_i$ is provided by the ETE, a ResNet-18~\cite{he2016deep} backbone pretrained via supervised contrastive learning~\cite{khosla2020supervised} to separate exposure anchors in a latent space. Conditioned on $\bm{u}_i$, a shallow network $\mathcal{M}$ predicts affine parameters $(\boldsymbol{\alpha}, \boldsymbol{\beta}) = \mathcal{M}(\bm{u}_i)$ and modulates the attention output $\hat{\bm{F}}_i^j$ as $\bm{F}_i^{j+1}=(1+\boldsymbol{\alpha})\odot \hat{\bm{F}}_i^j+\boldsymbol{\beta}$. This injects exposure information into all refinement stages with negligible overhead, enabling dynamic exposure adaptation.

In summary, compared to sliding windows~\cite{wang2019edvr, jo2018deep, li2020mucan, tao2017detail, Youk_2024_CVPR}, HRBA accesses long-range context via hierarchical aggregation. Compared to recurrent schemes~\cite{chan2021basicvsr, chan2022basicvsr++, liang2022recurrent, liu2022learning}, it avoids sequential dependencies, enabling efficient parallelization and stable training on long sequences.

\subsection{Degradation Learning and Restoration Networks} \label{sec:arch}
As outlined in Sec.~\ref{sec:overview}, our framework comprises two main networks leveraging the HRBA backbone to solve the inverse problem defined in Sec.~\ref{sec:formulation}.

\noindent\textbf{Degradation Learning Network (Net$^D$).}
As shown in the left part of Fig.~\ref{fig:network}, Net$^D$ aims to estimate degradation priors from the input blurry LR sequence $\bm{X}$. It processes $\bm{X}$ through a stack of HRBA blocks with integrated ETM layers, producing refined features $\bm{F}^{D,M}$ and multi-flow-mask pairs $\mathbf{f}^{D,M}$. From these outputs, Net$^D$ predicts two key priors for each frame $\bm{X}_i$: (i) image flow-mask pairs $\mathbf{f}^{\bm{Y}}_i = \bigl\{\bm{f}^{\bm{Y}}_{\,i \rightarrow i\pm 1},\,\bm{o}^{\bm{Y}}_{\,i \rightarrow i\pm 1}\bigr\}$, representing motion between the sharp HR frame $\bm{Y}_i$ and its neighbors, and (ii) degradation kernels $\mathcal{K}^{D}_i \in \mathbb{R}^{3 \times H \times W \times k_d^2}$.

To apply these predicted kernels, we utilize the Flow-Guided Dynamic Filtering (FGDF)~\cite{Youk_2024_CVPR} module. Crucially, because $\mathcal{K}^{D}_i$ is predicted from features already infused with exposure information via the ETM layer, this operation naturally becomes an exposure-aware FGDF. This kernel formulation, representing the degradation from three consecutive sharp HR frames $\{\bm{Y}_{i-1}, \bm{Y}_i, \bm{Y}_{i+1}\}$ to the blurry LR frame $\bm{X}_i$, follows the design principle of~\cite{Youk_2024_CVPR} as it offers a robust trade-off between performance and computational cost. The shape of $\mathcal{K}^{D}_i$ reflects its spatio-temporally variant and position-dependent nature, providing a structured learnable parameterization for modeling complex varying degradations.

To ensure accurate prior estimation, Net$^D$ is trained with a reconstruction objective: the predicted priors must reconstruct the blurry LR frame $\hat{\bm{X}}_i$ from the ground-truth (GT) sharp HR frames $\bm{Y}$ as:
\begin{equation}
   \hat{\bm{X}}_i = \left( \mathcal{K}^D_i \circledast_s \{\bm{Y}_{t \rightarrow i}\}_{t=i-1}^{i+1} \right),
   \label{eq:recon_xt}
\end{equation}
where $\circledast_s$ denotes the FGDF operation~\cite{Youk_2024_CVPR} with stride $s$, defined as:
\begin{equation}
\left(\mathcal{K}^D_i \circledast_s \{\bm{Y}_{t\rightarrow i}\}_{t=i-1}^{i+1}\right)(\bm{p})
=
\sum_{t=i-1}^{i+1}\;\sum_{k=1}^{k_d^2}
\mathcal{K}^D_{i,t}(\bm{p}, \bm{p}_k)\;
\bm{Y}_{t\rightarrow i}(s\bm{p}+\bm{p}_k),
\end{equation}
where $\bm{p}_k$ denotes the $k$-th spatial offset within the $k_d \times k_d$ grid. Unlike conventional dynamic filtering that uses fixed spatial neighborhoods, FGDF explicitly performs filtering \emph{along motion trajectories} by dynamically sampling features guided by the estimated optical flow. By utilizing our exposure-conditioned kernel $\mathcal{K}^D_i$, this operation naturally achieves jointly motion- and exposure-aware filtering, making Eq.~\ref{eq:recon_xt} a \textit{practical, learnable approximation} of the conceptual degradation model in Eq.~\ref{eq:conceptual_degradation}. The warped HR frame $\bm{Y}_{t \rightarrow i}$ is defined as:
\begin{equation}
   \bm{Y}_{t \rightarrow i} = 
   \begin{cases} 
       \bm{Y}_i, & \text{if } t = i \\ 
       \mathcal{W}(\bm{Y}_t, \bm{f}^{\bm{Y}}_{i \to t}, \bm{o}^{\bm{Y}}_{i \rightarrow t}), & \text{if } t = i\pm1
   \end{cases}
   \label{eq:warp}
\end{equation}
where $\mathcal{W}$ denotes the occlusion-aware backward warping~\cite{wolberg1990digital}.

\noindent\textbf{Restoration Network (Net$^R$).}
As shown in the right part of Fig.~\ref{fig:network}, Net$^R$ performs the final restoration, taking the blurry LR sequence $\bm{X}$ along with the rich priors predicted by Net$^D$ ($\bm{F}^{D,M}, \mathbf{f}^{D,M}$, and $\mathcal{K}^D$) as input. It first generates initial features by combining $\bm{X}$ and the context feature $\bm{F}^{D,M}$ using concatenation and an RDB~\cite{wang2018esrgan}. These features are then refined through another stack of HRBA blocks, initializing the multi-flow-mask pairs with $\mathbf{f}^{D,M}$ from Net$^D$ to leverage the motion prior. Crucially, within each HRBA block in Net$^R$, the DA attention utilizes the estimated kernel $\mathcal{K}^D_i$ as its query, after which the ETM layer continues to provide exposure conditioning, enabling degradation- and exposure-adaptive restoration. Finally, the refined features $\bm{F}^{R,M}$ pass through an upsampling block to predict a high-frequency residual $\hat{\bm{Y}}_i^{\text{res}}$. The final sharp HR frame is obtained by adding this residual to the bilinearly upsampled blurry LR input:
\begin{equation}
   \hat{\bm{Y}}_i = \hat{\bm{Y}}_i^{\text{res}} + \bm{X}_i\uparrow_s,
   \label{eq:final_output}
\end{equation}
where $\uparrow_s$ denotes the $\times s$ bilinear upsampling.

\begin{figure}[t]
\centering
\includegraphics[width=\linewidth]{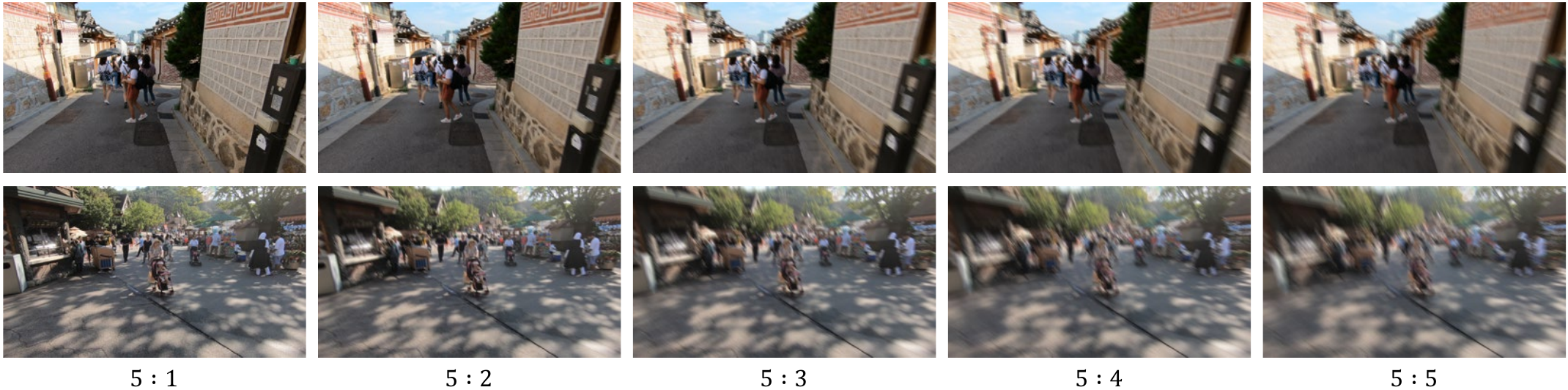}
\vspace{-6mm}
\caption{Example frames from our REDS-ME dataset across five exposure levels and two scenes. Each column corresponds to an exposure ratio from $5\!:\!1$ (shortest exposure) to $5\!:\!5$ (longest exposure). Longer exposures lead to increasingly severe motion blur, illustrating the controlled effect of exposure variation on blur extent.}
\label{fig:reds_me} 
\vspace{-6mm}
\end{figure}

\vspace{-2mm}
\subsection{Training Strategy}
\label{sec:strategy}
We adopt a three-stage training strategy to effectively optimize FMA-Net++. First, the ETE is pretrained using a supervised contrastive loss~\cite{khosla2020supervised} on exposure labels derived from our data synthesis process (Sec.~\ref{sec:datasets}; see \textit{Suppl.} for detailed contrastive loss formulations). Crucially, we freeze the ETE afterward to provide a stable exposure reference space, preventing representation drift during subsequent training. Second, guided by the frozen ETE, Net$^D$ is trained to predict reliable degradation priors, supervised by a composite loss $\mathcal{L}_D$:
\begin{equation}
\mathcal{L}_D = \sum_{i=1}^{T} l_1(\hat{\bm{X}}_i, \bm{X}_i) + \lambda_1 \sum_{i=1}^{T} l_1(\bm{Y}_{i \pm 1 \rightarrow i}, \bm{Y}_i) + \lambda_2 l_1(\bm{f}^{\bm{Y}}, \bm{f}^{\bm{Y}}_{\text{RAFT}}),
\label{eq:loss_d}
\end{equation}
where the first term is the reconstruction loss of the blurry LR input via Eq.~\ref{eq:recon_xt}, the second term is a warping loss, and the third term explicitly supervises the optical flow guided by a pretrained RAFT~\cite{teed2020raft}. RAFT is used only for training-time pseudo-supervision and is not required during inference.
Finally, the entire framework is jointly trained end-to-end, fine-tuning Net$^D$ alongside Net$^R$ with the total loss:
\begin{equation}
\mathcal{L}_{\mathrm{total}} = l_1(\hat{\bm{Y}}, \bm{Y}) + \lambda_3 \mathcal{L}_D,
\label{eq:loss_total}
\end{equation}
where the first term is the primary restoration loss on the final sharp HR output. 

\vspace{-2mm}
\section{Experiments}

\subsection{Experimental Setup} \label{sec:datasets}

\noindent\textbf{Datasets and Benchmarks.}
To systematically evaluate VSRDB under controlled exposure-duration variation, we construct two new benchmarks derived from the REDS dataset~\cite{nah2019ntire}: \textbf{REDS-ME} (Multi-Exposure) and \textbf{REDS-RE} (Random-Exposure).

For REDS-ME, we synthesize five distinct exposure levels corresponding to duty cycles from $5\!:\!1$ (shortest exposure) to $5\!:\!5$ (longest exposure), as illustrated in Fig.~\ref{fig:reds_me}. Crucially, these discrete levels serve as pseudo-labels for ETE pretraining. We train FMA-Net++ using all five levels from the REDS-ME training set and evaluate on the most challenging levels ($5\!:\!4$, $5\!:\!5$) of the REDS4-ME test set (derived from the standard REDS4 partition). To explicitly assess robustness to dynamic exposure variations, we construct REDS-RE by temporally mixing frames across the five exposure levels within the REDS4-ME test scenes, yielding controlled exposure trajectories with varying blur extents. Furthermore, we use the standard GoPro dataset~\cite{nah2017deep} to evaluate generalization to different scene and motion statistics, alongside qualitative assessments on real-world blurry videos. Details of the data synthesis pipeline are provided in the \textit{Suppl}.

\noindent\textbf{Evaluation Metrics.}
We evaluate restoration quality using PSNR and SSIM~\cite{wang2004image}. Temporal consistency is measured by tOF~\cite{chu2020learning}. For real-world videos where GT is unavailable, we report no-reference metrics such as NIQE~\cite{mittal2012making} and MUSIQ~\cite{ke2021musiq}. We also compare model efficiency in terms of parameter count and runtime.

\noindent\textbf{Implementation Details.}
All implementation details, including network configurations and hyperparameter settings, are provided in the \textit{Suppl.}

\renewcommand{\arraystretch}{0.9}
\begin{table*}[t]
\begin{center}
\caption{Quantitative comparison of $\times 4$ VSRDB on REDS4-ME for two challenging exposure levels ($5\!:\!4$ and $5\!:\!5$). All metrics are computed on the RGB channels. {\color{red}\textbf{Red}} and {\color{blue}blue} indicate the best and second-best performance, respectively. Runtime is measured per LR frame of resolution $180 \times 320$. The superscript $^*$ denotes models retrained on our proposed REDS-ME training set.}
\vspace{-5mm}
\label{tab:reds_me4_comparison}
\resizebox{\textwidth}{!}{
\begin{tabular}{c | c | c | c c c | c c c}
\hline
\multicolumn{1}{c|}{\multirow{2}{*}{Methods}} & \multirow{2}{*}{\# Params (M)} & \multirow{2}{*}{Runtime (s)} & \multicolumn{3}{c|}{REDS4-ME-$5\!:\!4$} & \multicolumn{3}{c}{REDS4-ME-$5\!:\!5$} \\
\cline{4-9}
\multicolumn{1}{c|}{} & & & PSNR$\uparrow$ & SSIM$\uparrow$ & tOF$\downarrow$ & PSNR$\uparrow$ & SSIM$\uparrow$ & tOF$\downarrow$ \\
\hline
\multicolumn{9}{c}{Super-Resolution + Deblurring} \\
\hline
SwinIR \cite{liang2021swinir} + Restormer \cite{zamir2022restormer} & 11.9 + 26.1 & 0.221 + 0.753 & 26.23 & 0.7464 & 3.775 & 25.53 & 0.7229 & 4.558 \\
HAT \cite{chen2023activating} + FFTformer \cite{kong2023efficient}  & 20.8 + 16.6 & 0.352 + 1.414 & 26.66 & 0.7634 & 3.207 & 25.92 & 0.7400 & 3.995 \\
BasicVSR++ \cite{chan2022basicvsr++} + RVRT \cite{liang2022recurrent}      & 7.3 + 13.6  & 0.048 + 0.349 & 27.28 & 0.7901 & 2.887 & 26.98 & 0.7621 & 3.164 \\
IART \cite{xu2024enhancing} + BSSTNet \cite{zhang2024blur}                & 13.4 + 52.0 & 1.041 + 0.482 & 27.50 & 0.8006 & 2.578 & 27.26 & 0.7888 & 2.721 \\
\hline
\multicolumn{9}{c}{Deblurring + Super-Resolution} \\
\hline
Restormer \cite{zamir2022restormer} + SwinIR \cite{liang2021swinir}       & 26.1 + 11.9 & 0.043 + 0.221 & 26.36 & 0.7499 & 3.464 & 25.84 & 0.7316 & 3.948 \\
FFTformer \cite{kong2023efficient} + HAT \cite{chen2023activating}        & 16.6 + 20.8 & 0.066 + 0.352 & 26.36 & 0.7534 & 3.256 & 25.87 & 0.7356 & 3.739 \\
RVRT \cite{liang2022recurrent} + BasicVSR++ \cite{chan2022basicvsr++}     & 13.6 + 7.3  & 0.019 + 0.048 & 26.35 & 0.7492 & 3.314 & 25.95 & 0.7424 & 3.610 \\
BSSTNet \cite{zhang2024blur} + IART \cite{xu2024enhancing}                & 52.0 + 13.4 & 0.025 + 1.041 & 26.51 & 0.7711 & 3.103 & 26.33 & 0.7564 & 3.313 \\
\hline
\multicolumn{9}{c}{Blind Video Super-Resolution} \\
\hline
DBVSR \cite{pan2021deep} & 14.1 & 0.096 & 24.50 & 0.7208 & 3.449 & 22.19 & 0.6122 & 4.554 \\
\hline
\multicolumn{9}{c}{Joint Video Super-Resolution and Deblurring} \\
\hline
Restormer$^*$ \cite{zamir2022restormer}      & 26.5 & 0.045 & 27.45 & 0.7851 & 2.161 & 27.12 & 0.7750 & 2.516 \\
DBVSR$^*$ \cite{pan2021deep}                 & 14.1 & 0.096 & 26.77 & 0.7629 & 3.021 & 26.07 & 0.7405 & 3.765 \\
BasicVSR++$^*$ \cite{chan2022basicvsr++}     & 7.3  & 0.048 & 27.70 & 0.7922 & 2.302 & 27.14 & 0.7770 & 2.746 \\
IART$^*$ \cite{xu2024enhancing}              & 13.4 & 1.041 & 28.23 & 0.8153 & 2.143 & 27.64 & 0.7972 & 2.590 \\
VRT$^*$ \cite{liang2024vrt}           & 35.6 & 0.684 & 27.93 & 0.8045 & 2.366 & 27.41 & 0.7887 & 2.839 \\
RVRT$^*$ \cite{liang2022recurrent}           & 12.9 & 0.385 & 28.11 & 0.8093 & 2.136 & 27.58 & 0.7944 & 2.558 \\
BSSTNet$^*$ \cite{zhang2024blur}             & 52.0 & 0.548 & 28.75 & {\color{blue}0.8342} & 1.893 & 28.11 & 0.8119 & 2.298 \\
Ev-DeblurVSR \cite{kai2025event}             & 8.3  & 0.062 & 24.51 & 0.7154 & 3.602 & 24.38 & 0.7047 & 4.094 \\
Ev-DeblurVSR$^*$ \cite{kai2025event}         & 8.3  & 0.062 & 27.40 & 0.7839 & 2.521 & 26.82 & 0.7672 & 3.059 \\
FMA-Net \cite{Youk_2024_CVPR}                & 9.6  & 0.318 & 26.42 & 0.7958 & 2.503 & 26.67 & 0.8005 & 2.443 \\
FMA-Net$^*$ \cite{Youk_2024_CVPR}            & 9.6  & 0.318 & {\color{blue}29.04} & 0.8275 & {\color{blue}1.891} & {\color{blue}28.51} & {\color{blue}0.8136} & {\color{blue}2.269} \\
\hline
FMA-Net++ (Ours)                             & 12.8 & 0.074 & {\color{red}\textbf{29.66}} & {\color{red}\textbf{0.8546}} & {\color{red}\textbf{1.688}} & {\color{red}\textbf{29.24}} & {\color{red}\textbf{0.8453}} & {\color{red}\textbf{1.956}} \\
\hline
\end{tabular}}
\end{center}
\vspace{-10mm}
\end{table*}

\subsection{Comparisons with State-of-the-Art Methods}
We compare FMA-Net++ against SOTA methods across relevant categories: single-image SR (SwinIR~\cite{liang2021swinir}, HAT~\cite{chen2023activating}), single-image deblurring (Restormer~\cite{zamir2022restormer}, FFTformer~\cite{kong2023efficient}), VSR (BasicVSR++~\cite{chan2022basicvsr++}, IART~\cite{xu2024enhancing}), video deblurring (VRT~\cite{liang2024vrt}, RVRT~\cite{liang2022recurrent}, BSSTNet~\cite{zhang2024blur}), Blind VSR (DBVSR~\cite{pan2021deep}), and joint VSRDB (FMA-Net~\cite{Youk_2024_CVPR}, Ev-DeblurVSR~\cite{kai2025event}). For a fair comparison in the joint VSRDB setting under varying exposure conditions, relevant SOTA methods were adapted and retrained on our REDS-ME training set, denoted by $^*$ in Tables \ref{tab:reds_me4_comparison} and \ref{tab:gopro_comparison}.

\noindent\textbf{Quantitative Results.}
Table~\ref{tab:reds_me4_comparison} presents the performance on REDS4-ME across two challenging exposure levels ($5\!:\!4$ and $5\!:\!5$), representing severe motion blur. FMA-Net++ consistently outperforms all baselines across PSNR, SSIM, and tOF. For instance, FMA-Net++ achieves significant gains of 0.62 dB and 0.73 dB over the second-best model, FMA-Net$^*$, on levels $5\!:\!4$ and $5\!:\!5$, respectively. Furthermore, FMA-Net++ demonstrates superior efficiency compared to methods with similar complexity like RVRT$^*$~\cite{liang2022recurrent}. It achieves remarkably higher performance while being significantly faster (over 5.2$\times$ speedup), an efficiency that primarily arises from our parallelizable HRBA architecture. While VRT~\cite{liang2024vrt} also enables sequence-level parallelization, it suffers from massive memory consumption and a heavy computational burden, requiring 20.5 GB for a 10-frame sequence at $180 \times 320$ resolution. In contrast, our FMA-Net++ requires only 6.2 GB, demonstrating that HRBA achieves parallel temporal modeling far more efficiently while maintaining state-of-the-art accuracy.

Table~\ref{tab:gopro_comparison} evaluates robustness to dynamic exposure (REDS-RE) and generalization ability to an unseen dataset (GoPro~\cite{nah2017deep}). On REDS-RE, featuring dynamic exposure transitions within sequences, the performance advantage of FMA-Net++ over other methods widens considerably compared to REDS-ME. This result validates the effectiveness of our explicit exposure-aware modeling (ETM) in adapting to temporally varying exposure conditions where fixed-exposure assumptions struggle. On the unseen GoPro dataset, which has different scene and motion statistics from REDS-ME, FMA-Net++ again achieves the best performance across all metrics, indicating strong generalization beyond the training domain.

\renewcommand{\arraystretch}{0.95}
\begin{table}[b]
\begin{center}
\vspace{-5mm}
\caption{Quantitative comparison of $\times 4$ VSRDB on REDS-RE and GoPro \cite{nah2017deep}.}
\vspace{-3mm}
\label{tab:gopro_comparison}
\scalebox{0.8}{
\begin{tabular}{c | c c c | c c c}
\hline
\multicolumn{1}{c|}{\multirow{2}{*}{Methods}} & \multicolumn{3}{c|}{REDS-RE} & \multicolumn{3}{c}{GoPro} \\
\cline{2-7}
\multicolumn{1}{c|}{} & PSNR $\uparrow$ & SSIM $\uparrow$ & tOF $\downarrow$ & PSNR $\uparrow$ & SSIM $\uparrow$ & tOF $\downarrow$ \\ 
\hline
Restormer$^*$ \cite{zamir2022restormer}       & 27.79 & 0.7953 & 1.775 & 27.54 & 0.8350 & 3.302 \\
DBVSR$^*$ \cite{pan2021deep}                 & 27.30 & 0.7742 & 2.398 & 26.05 & 0.7815 & 4.730 \\
BasicVSR++$^*$ \cite{chan2022basicvsr++}     & 28.14 & 0.8044 & 1.904 & 27.40 & 0.8282 & 3.285 \\
IART$^*$ \cite{xu2024enhancing}              & 28.68 & 0.8248 & 1.852 & 27.76 & 0.8394 & 3.302 \\
VRT$^*$ \cite{liang2024vrt}                           & 28.24     & 0.8124      & 2.071     & 27.39     & 0.8304      & 3.616     \\
RVRT$^*$ \cite{liang2022recurrent}           & 28.56 & 0.8208 & 1.926 & 27.64 & 0.8364 & 3.223 \\
BSSTNet$^*$ \cite{zhang2024blur}             & {\color{blue}29.33} & {\color{blue}0.8427} & {\color{blue}1.602} & 28.57 & 0.8650 & 2.753 \\
Ev-DeblurVSR$^*$ \cite{kai2025event}         & 27.94 & 0.7987 & 2.039 & 27.25 & 0.8247 & 3.536 \\
FMA-Net$^*$ \cite{Youk_2024_CVPR}            & 29.29 & 0.8413 & 1.614 & {\color{blue}28.83} & {\color{blue}0.8655} & {\color{blue}2.727} \\ 
\hline
FMA-Net++ (Ours)                             & {\color{red}\textbf{30.13}} & {\color{red}\textbf{0.8643}} & {\color{red}\textbf{1.360}} & {\color{red}\textbf{30.49}} & {\color{red}\textbf{0.9018}} & {\color{red}\textbf{2.091}} \\ 
\hline
\end{tabular}}
\end{center}
\vspace{-8mm}
\end{table}

\begin{figure}[t]
\centering
\includegraphics[width=\linewidth]{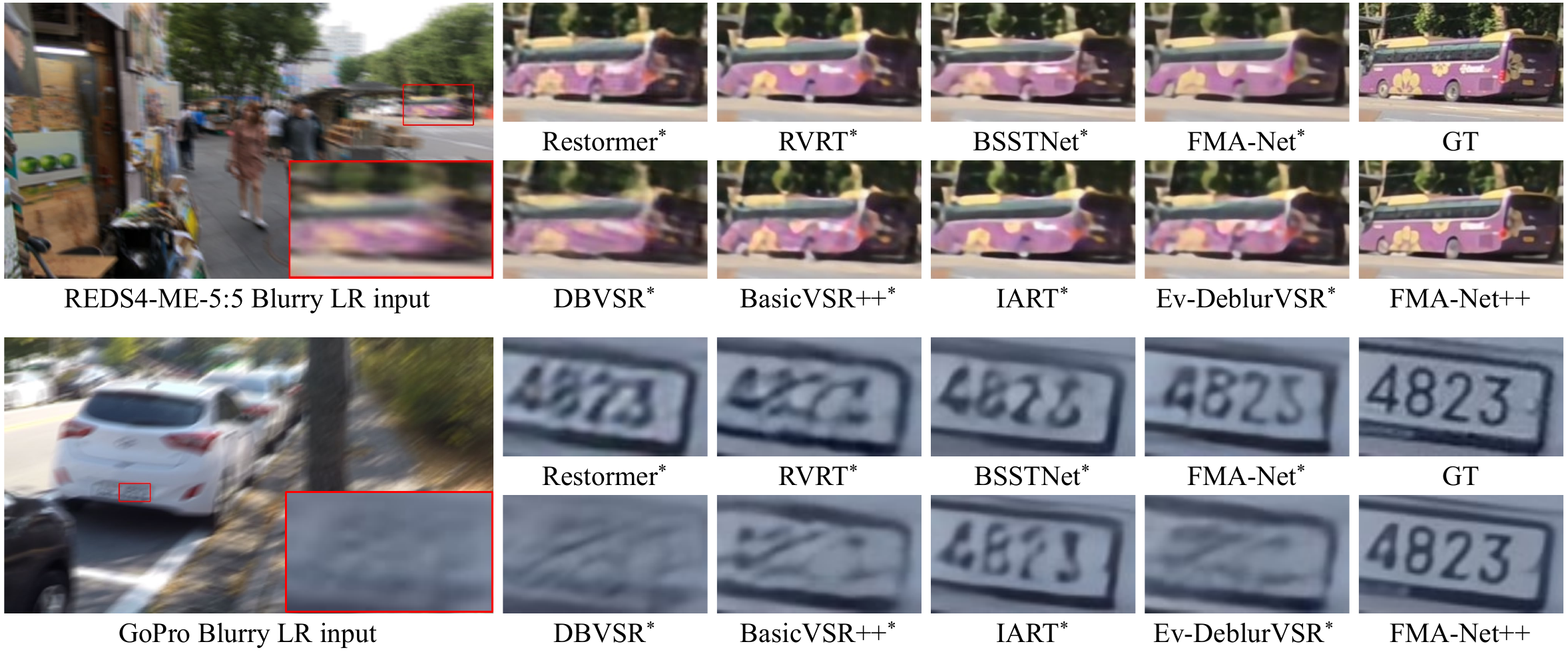}
\vspace{-4mm}
\caption{Qualitative comparisons of $\times 4$ VSRDB on REDS4-ME-$5\!:\!5$ and GoPro~\cite{nah2017deep}. Each scene contains severe motion blur with different characteristics.}
\label{fig:qualitative_comparison}
\vspace{-5mm}
\end{figure}

\noindent\textbf{Qualitative Results.}
Fig.~\ref{fig:qualitative_comparison} presents visual comparisons on synthetic benchmarks (REDS4-ME-$5\!:\!5$ and GoPro) that contain severe motion blur, while Fig.~\ref{fig:teaser}(a) shows results on challenging real-world videos captured with a smartphone. On both synthetic and real-world data, FMA-Net++ consistently restores sharper details, cleaner edges, and more legible text with fewer artifacts, achieving the best perceptual quality (NIQE/MUSIQ). We omit multi-modal methods such as Ev-DeblurVSR~\cite{kai2025event} from the real-world comparison, as they are fundamentally not applicable to standard RGB videos that lack the required event data. This highlights the strong practicality and generalization of our approach, which achieves these results using only conventional RGB inputs despite being trained solely on synthetic data. Further qualitative results are provided in the \textit{Suppl}.

\renewcommand{\arraystretch}{0.9}
\begin{table}[b]
\begin{center}
\vspace{-4mm}
\caption{Comparison of temporal modeling variants within FMA-Net++ on REDS4-ME-$5\!:\!5$. Our HRBA design achieves the best accuracy and temporal consistency.}
\vspace{-2mm}
\label{tab:abla_arch_comp}
\scalebox{0.9}{
\begin{tabular}{l|c|c|c}
\hline
\multicolumn{1}{c|}{Temporal Modeling Strategy} & Runtime (s)  & PSNR $\uparrow$  & tOF $\downarrow$ \\
\hline
Sliding-window variant & 0.314 & 28.57 & 2.231 \\
Recurrent propagation variant & 0.086 & 29.11 & 1.989 \\
\textbf{HRBA (Ours)} & \textbf{0.074} & \textbf{29.24} & \textbf{1.956} \\
\hline
\end{tabular}}
\end{center}
\vspace{-8mm}
\end{table}

\renewcommand{\arraystretch}{1.0}
\begin{table*}[t]
\begin{center}
\caption{Ablation study on the Exposure Time-aware Feature Extractor (ETE) for multiple datasets.}
\vspace{-5mm}
\label{tab:abla_ete}
\resizebox{\textwidth}{!}{
\begin{tabular}{l | c | c | c c | c c | c c | c c}
\hline
\multicolumn{1}{c|}{\multirow{3}{*}{Methods}} & \multirow{3}{*}{\shortstack{\# Params\\(M)}} & \multirow{3}{*}{\shortstack{Runtime\\(s)}} & \multicolumn{4}{c|}{In-distribution} & \multicolumn{4}{c}{Out-of-distribution} \\
\cline{4-11}
\multicolumn{1}{c|}{} & & & \multicolumn{2}{c|}{REDS4-ME-$5\!:\!4$} & \multicolumn{2}{c|}{REDS4-ME-$5\!:\!5$} & \multicolumn{2}{c|}{REDS-RE} & \multicolumn{2}{c}{GoPro} \\
\cline{4-11}
\multicolumn{1}{c|}{} & & & PSNR$\uparrow$ & tOF$\downarrow$ & PSNR$\uparrow$ & tOF$\downarrow$ & PSNR$\uparrow$ & tOF$\downarrow$ & PSNR$\uparrow$ & tOF$\downarrow$ \\
\hline
FMA-Net++ w/o ETE & 9.8  & 0.071 & 29.55 & 1.764 & 29.12 & 2.054 & 29.72 & 1.436 & 29.78 & 2.267 \\
FMA-Net++ w/o ETE (matched) & 13.1 & 0.088 & \textbf{29.67} & 1.713 & 29.20 & 2.011 & 29.88 & 1.399 & 29.85 & 2.251 \\
FMA-Net++ w/ ETE  & 12.8 & 0.074 & 29.66 & \textbf{1.688} & \textbf{29.24} & \textbf{1.956} & \textbf{30.13} & \textbf{1.360} & \textbf{30.49} & \textbf{2.091} \\
\hline
\end{tabular}}
\end{center}
\vspace{-8mm}
\end{table*}

\vspace{-2mm}
\section{Ablation Study} \label{sec:ablation}
We present ablation studies validating our key design choices. Further ablation studies and detailed analyses can be found in the \textit{Suppl}.

\vspace{-2mm}
\subsection{Effectiveness of Hierarchical Architecture}
To validate the advantages of our proposed hierarchical temporal architecture (conceptually compared with other strategies in Fig.~\ref{fig:concept}(a)), we compare the full FMA-Net++ against two variants built upon its core components but employing different temporal modeling strategies: (i) a sliding-window variant processing three frames at a time, similar to~\cite{Youk_2024_CVPR}, and (ii) a recurrent variant where the hierarchical refinement is adapted for sequential propagation. All variants maintain the same number of HRBA blocks and utilize the same ETM and multi-attention mechanisms, forming an identical backbone for a fair comparison.

Table~\ref{tab:abla_arch_comp} presents the comparison results on REDS4-ME-$5\!:\!5$. Our hierarchical FMA-Net++ demonstrates substantial improvements over both variants. Compared to the sliding-window variant, it yields markedly better results across all metrics, effectively overcoming the limitations imposed by a fixed temporal receptive field. Compared to the recurrent variant, it achieves superior performance in both PSNR and tOF. This noticeable gain in temporal consistency likely stems from its non-recurrent hierarchical structure, which mitigates gradient-vanishing issues that can affect sequential propagation over long sequences. We also empirically observe that this design achieves the most stable training dynamics among the compared variants. Furthermore, in terms of efficiency, the hierarchical design demonstrates a modest speed advantage over the recurrent approach. Overall, these results validate that our hierarchical strategy serves as a highly effective backbone for high-quality and temporally consistent video restoration.

\begin{figure}[h]
\vspace{-7mm}
\begin{minipage}[b]{0.45\linewidth}
    \centering
    \includegraphics[width=0.6\linewidth]{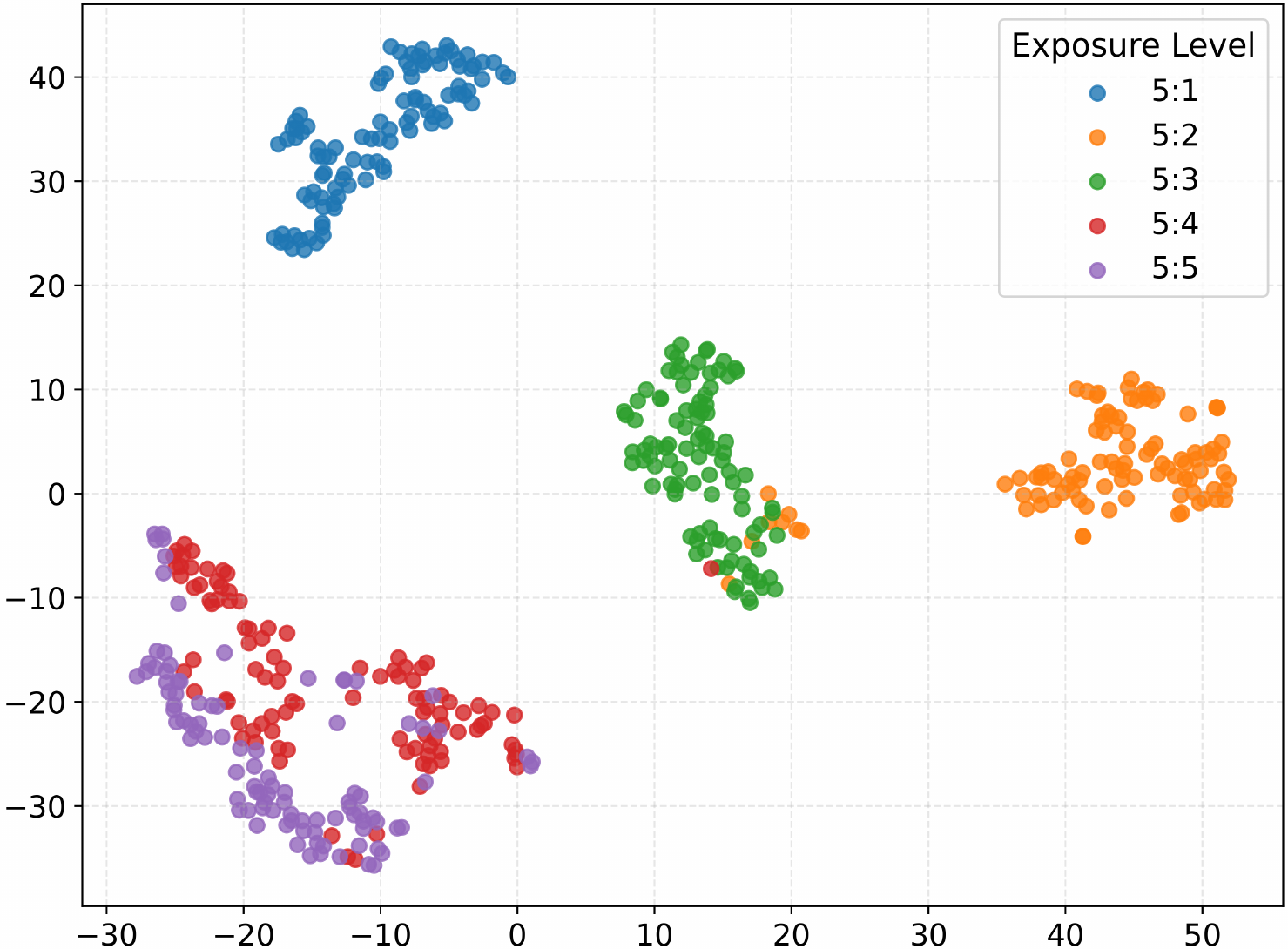}
    \vspace{-2mm}
    \centerline{(a) t-SNE visualization of ETE features}
\end{minipage}
\hfill
\begin{minipage}[b]{0.53\linewidth}
    \centering
    \includegraphics[width=\linewidth]{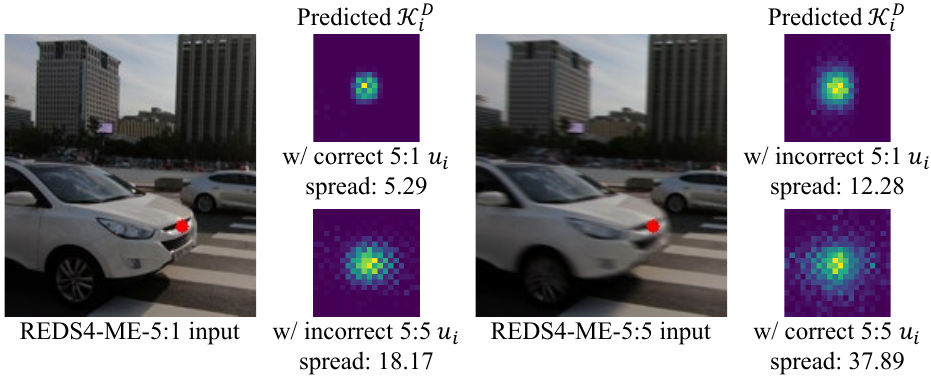}
    \vspace{-2mm}
    \centerline{(b) Effect of guidance on kernels}
\end{minipage}
\caption{Visual analysis of exposure-aware modeling. (a) t-SNE~\cite{van2008visualizing} visualization of ETE features ($\bm{u}$) shows clear clustering across exposure levels. (b) Bidirectional guidance test: for both a $5\!:\!5$ and a $5\!:\!1$ input, the predicted degradation kernel adapts its spatial spread to the supplied exposure guidance.}
\label{fig:visual_analysis}
\vspace{-8mm}
\end{figure}

\vspace{-4mm}
\subsection{Effectiveness of Exposure-Aware Modeling} \label{sec:abla_exposure}
\vspace{-2mm}
Table~\ref{tab:abla_ete} evaluates our exposure-aware modeling by comparing FMA-Net++ with variants without ETE, including a capacity-matched baseline (13.1M), slightly larger than the full model (12.8M). Increasing the capacity of the w/o-ETE model absorbs most of the in-distribution PSNR gain on REDS4-ME, indicating that part of the improvement under fixed exposure comes from backbone capacity. However, the matched baseline does not close the gap on the exposure-varying REDS-RE split, where FMA-Net++ w/ ETE still improves PSNR from 29.88 to 30.13 and reduces tOF from 1.399 to 1.360. This indicates that the ETE/ETM pipeline contributes to exposure-conditioned robustness beyond parameter count. The consistent gain on the unseen GoPro dataset (+0.64 dB) further reflects improved generalization beyond the training domain.

\noindent\textbf{Choice of ETE Design.}
To justify the contrastive ETE design, we compare it with three deployable exposure-conditioning alternatives that compute guidance from the input: frame-difference features, a classification-pretrained ETE, and an ordinal-regression ETE. As shown in Table~\ref{tab:ete_design}, all alternatives underperform our contrastive ETE on REDS-RE, indicating that the gain does not come from merely adding an arbitrary exposure-related signal. The contrastive objective provides a more effective exposure embedding for the ETM/FGDF pipeline.

\noindent \textbf{Visual Analysis of Learned Priors.}
To further validate this explicit conditioning, we visualize the learned representations in Fig.~\ref{fig:visual_analysis}. The t-SNE~\cite{van2008visualizing} visualization of ETE features ($\bm{u}$) in Fig.~\ref{fig:visual_analysis}(a) shows clear clustering across exposure levels, confirming that ETE successfully extracts exposure-dependent characteristics. Furthermore, Fig.~\ref{fig:visual_analysis}(b) demonstrates the impact of this guidance through a bidirectional test. For a severely blurred $5\!:\!5$ input, correct $5\!:\!5$ guidance yields a spatially diffuse kernel, whereas incorrect $5\!:\!1$ guidance contracts it into a concentrated one. Conversely, for a mildly blurred $5\!:\!1$ input, $5\!:\!5$ guidance drives the kernel to become more diffuse. These bidirectional responses show that the predicted kernel $\mathcal{K}^D_i$ adapts its spatial shape to the supplied exposure guidance.

\renewcommand{\arraystretch}{0.95}
\begin{table}[b]
\vspace{-4mm}
\begin{minipage}[t]{0.47\linewidth}
    \centering
    \caption{Comparison of exposure-conditioning designs on REDS-RE.}
    \vspace{-2mm}
    \label{tab:ete_design}
    \resizebox{\linewidth}{!}{
    \begin{tabular}{l|c}
    \hline
    Conditioning design & PSNR$\uparrow$ / tOF$\downarrow$ \\
    \hline
    Frame-difference            & 29.75 / 1.429 \\
    Classification-pretrained ETE          & 29.92 / 1.398 \\
    Ordinal-regression ETE      & 29.97 / 1.382 \\
    Contrastive ETE (Ours)      & \textbf{30.13} / \textbf{1.360} \\
    \hline
    \end{tabular}}
\end{minipage}
\hfill
\begin{minipage}[t]{0.50\linewidth}
    \centering
    \caption{Effect of corrupting frame-wise guidance $\bm{u}$ on REDS-RE.}
    \vspace{-2mm}
    \label{tab:wrong_guidance}
    \resizebox{\linewidth}{!}{
    \begin{tabular}{l|c|c}
    \hline
    \multirow{2}{*}{Guidance $\bm{u}$} & Average & Transition Frames \\
    \cline{2-3}
    & PSNR$\uparrow$ / tOF$\downarrow$ & PSNR$\uparrow$ / tOF$\downarrow$ \\
    \hline
    Correct                     & \textbf{30.13} / \textbf{1.360} & \textbf{29.98} / \textbf{1.395} \\
    Same-frame fixed $5\!:\!5$  & 29.88 / 1.387 & 29.40 / 1.631 \\
    Sequence-shuffled           & 29.59 / 1.525 & 29.18 / 1.713 \\
    Random wrong-level          & 29.47 / 1.605 & 29.05 / 1.739 \\
    w/o ETE (no guidance)      & 29.72 / 1.436 & 29.48 / 1.633 \\
    \hline
    \end{tabular}}
\end{minipage}
\end{table}

\noindent\textbf{Frame-wise Guidance Analysis.}
We further analyze the role of ETE guidance on REDS-RE by keeping the input sequence fixed and corrupting only the frame-wise guidance $\bm{u}$, as shown in Table~\ref{tab:wrong_guidance}. Correct guidance performs best, while fixed, sequence-shuffled, and random wrong-level guidance degrade both PSNR and tOF, with larger drops on transition frames where the exposure level changes. Notably, random wrong-level guidance performs worse than removing guidance entirely, showing that $\bm{u}$ is actively used rather than ignored.

\noindent \textbf{Fixed-Input Sensitivity.} 
Complementary to the REDS-RE corruption test, Table~\ref{tab:ete_sensitivity} reports the effect of replacing $\bm{u}$ with exposure embeddings from different levels on fixed REDS4-ME-$5\!:\!5$ inputs. Performance decreases gradually as the guidance deviates from the correct level, yet remains stable even under the farthest $5\!:\!1$ guidance. This shows that, when the input exposure is fixed and strong spatio-temporal evidence is available, FMA-Net++ leverages its HRBA backbone rather than relying exclusively on ETE.

\vspace{-4mm}
\subsection{Effectiveness of HRBA and Core Components} \label{sec:ablation_comp}
We conduct ablation studies to validate the key components and design choices of FMA-Net++, summarizing the main results in Table~\ref{tab:abla_arch}.

First, we investigate the impact of our hierarchical refinement strategy by varying the number of stacked HRBA blocks ($M$). As shown in Table~\ref{tab:abla_arch}(a, b, d), increasing $M$ from 1 to 2 and finally to our full configuration ($M=4$, row d) progressively improves both PSNR and tOF. This demonstrates the effectiveness of hierarchically expanding the temporal receptive field. As visualized in the \textit{Suppl.}, features become progressively sharper and more structurally aligned through the stacked blocks, further validating our hierarchical design.

Next, we validate the effectiveness of the Degradation-Aware (DA) attention within Net$^R$'s multi-attention module. Replacing DA attention with a standard SFT layer~\cite{wang2018recovering} for modulation significantly degrades performance, confirming that explicitly leveraging the estimated degradation priors via DA attention is crucial for targeted restoration.

\renewcommand{\arraystretch}{0.95}
\begin{table}[t]
\begin{minipage}[t]{0.47\linewidth}
    \centering
    \caption{Sensitivity to ETE guidance on fixed REDS4-ME-$5\!:\!5$ inputs.}
    \vspace{-2mm}
    \label{tab:ete_sensitivity}
    \resizebox{\linewidth}{!}{
    \begin{tabular}{c|c|c}
    \hline
    Input Frame & Guidance $\bm{u}$ & PSNR $\uparrow$ / tOF $\downarrow$ \\
    \hline
    \multirow{5}{*}{$5\!:\!5$} 
    & from $5\!:\!5$ (Correct) & \textbf{29.24} / \textbf{1.956} \\
    & from $5\!:\!4$           & 29.20 / 1.972 \\
    & from $5\!:\!3$           & 29.13 / 2.012 \\
    & from $5\!:\!2$           & 29.11 / 2.027 \\
    & from $5\!:\!1$           & 29.07 / 2.041 \\
    \hline
    \multicolumn{2}{c|}{Baseline w/o ETE} & 29.12 / 2.054 \\
    \hline
    \end{tabular}}
\end{minipage}
\hfill
\begin{minipage}[t]{0.49\linewidth}
    \centering
    \caption{Ablation on key components of FMA-Net++.}
    \vspace{-2mm}
    \label{tab:abla_arch}
    \resizebox{\linewidth}{!}{
    \begin{tabular}{l | c | c | c c c}
    \hline
    \multicolumn{1}{c|}{Methods} & \# Params & Time (s) & PSNR$\uparrow$ & SSIM$\uparrow$ & tOF$\downarrow$ \\
    \hline
    \multicolumn{6}{c}{The number of HRBA blocks $M$} \\
    \hline
    (a) $M=1$ & 7.7M & 0.035 & 28.29 & 0.8174 & 2.461 \\
    (b) $M=2$ & 9.4M & 0.048 & 28.74 & 0.8339 & 2.151 \\
    \hline
    \multicolumn{6}{c}{Multi-Attention} \\
    \hline
    (c) self-attn + SFT & 13.2M & 0.066 & 28.86 & 0.8378 & 2.132 \\
    \hline
    \multicolumn{6}{c}{Proposed} \\
    \hline
    (d) FMA-Net++ & 12.8M & 0.074 & \textbf{29.24} & \textbf{0.8453} & \textbf{1.956} \\
    \hline
    \end{tabular}}
\end{minipage}
\vspace{-5mm}
\end{table}

\vspace{-2mm}
\section{Conclusion}
\vspace{-2mm}
In this paper, we addressed the challenging problem of joint VSRDB under unknown and dynamically varying exposure conditions. To tackle this challenge, we introduced FMA-Net++, a novel framework built upon HRBA blocks that enables effective sequence-level temporal modeling with efficient parallel processing. Crucially, our proposed ETM layer injects per-frame exposure conditioning into the features. This allows our exposure-aware FGDF module to predict degradation kernels that capture the coupled effects of motion and exposure. Extensive experiments on the proposed REDS-ME and REDS-RE benchmarks, as well as GoPro and real-world videos, demonstrate that FMA-Net++ achieves state-of-the-art results, showcasing superior performance, efficiency, and robustness while generalizing effectively despite being trained on synthetic data.

\vspace{-2mm}
\subsubsection*{Acknowledgements.}
This work was supported by Institute of Information \& communications Technology Planning \& Evaluation (IITP) grant funded by the Korean Government [Ministry of Science and ICT (Information and Communications Technology)] (Project Number: RS-2022-00144444, Project Title: Deep Learning Based Visual Representational Learning and Rendering of Static and Dynamic Scenes).

\clearpage
\bibliographystyle{splncs04}
\bibliography{main}

\clearpage

\renewcommand{\thefigure}{S\arabic{figure}}
\renewcommand{\thetable}{S\arabic{table}}
\renewcommand{\theequation}{S\arabic{equation}}
\renewcommand{\thealgorithm}{S\arabic{algorithm}}
\renewcommand{\thesection}{S\arabic{section}}
\renewcommand{\thesubsection}{S\arabic{section}.\arabic{subsection}}
\setcounter{figure}{0}
\setcounter{table}{0}
\setcounter{equation}{0}
\setcounter{section}{0}

\title{FMA-Net++: Supplementary Material}
\author{}
\institute{}
\maketitle
\begin{sloppypar}
In this \textit{Supplementary Material}, we provide additional details and results to support the main paper. First, we present the detailed training strategy (Sec.~\ref{sec:supp_strategy}) and comprehensive implementation details (Sec.~\ref{sec:supp_impl}). Next, we detail the synthesis process for our proposed REDS-ME and REDS-RE benchmarks (Sec.~\ref{sec:supp_datasets}). Furthermore, we present further ablation studies and analyses (Sec.~\ref{sec:supp_abla}). Finally, we provide additional qualitative comparisons (Sec.~\ref{sec:supp_qual}), and discuss the limitations and future directions of our FMA-Net++ (Sec.~\ref{sec:supp_limit}).

\section{Detailed Training Strategy} \label{sec:supp_strategy}
As described in Sec.~\ref{sec:strategy} of the main paper, we adopt a three-stage training strategy where the Exposure Time-aware Feature Extractor (ETE) is first pretrained and subsequently frozen. In this first stage, the ETE is trained to learn an exposure-discriminative representation space. To achieve this, we apply supervised contrastive learning~\cite{khosla2020supervised}, utilizing the discrete exposure levels from our synthetic dataset (detailed in Sec.~\ref{sec:datasets} of the main paper and Sec.~\ref{sec:supp_datasets}) as pseudo-labels. The ETE is optimized to minimize the following contrastive loss $\mathcal{L}_{e}$:
\begin{equation}
    \mathcal{L}_{e} = \sum_{\bm{q} \in \mathcal{B}} - \frac{1}{|\mathcal{P}|} \sum_{\bm{p} \in \mathcal{P}} \log \frac{\exp \left(\bm{q}^\top \bm{p} / \alpha \right)}{\sum\limits_{\bm{p}' \in \mathcal{B} \setminus \{\bm{q}\}} \exp \left(\bm{q}^\top \bm{p}' / \alpha \right)},
\label{eq:supcon}
\end{equation}
where $\bm{q}$ denotes the anchor feature extracted by the ETE, $\mathcal{P}$ contains positive samples within the mini-batch $\mathcal{B}$ that share the same exposure label as the anchor, and $\alpha$ is a temperature scaling parameter. 

By freezing the ETE after this pretraining stage, we establish a stable exposure reference space. This design effectively prevents representation drift when subsequently optimizing Net$^D$ and Net$^R$ with $\mathcal{L}_D$ and $\mathcal{L}_{\mathrm{total}}$, which are defined in Eqs.~\ref{eq:loss_d} and~\ref{eq:loss_total} of the main paper.

\section{Implementation Details} \label{sec:supp_impl}
We train FMA-Net++ using the Adam optimizer~\cite{kingma2014adam} with default settings on 4 NVIDIA A6000 GPUs. In the first training stage, the ETE is trained with a mini-batch size of $128$, a learning rate of $0.01$, and a temperature scaling parameter $\alpha=0.5$ (as defined in Eq.~\ref{eq:supcon}). In the second stage, Net$^D$ is trained with a mini-batch size of $8$, using an initial learning rate of $2\times10^{-4}$ that is reduced by half at $70\%$, $85\%$, and $95\%$ of the total $280\text{K}$ iterations. The third stage jointly trains both Net$^D$ and Net$^R$ with the same batch size and learning rate schedule as in the second stage. 

FMA-Net++ is trained on 10-frame input sequences with a spatial patch size of $64 \times 64$ and evaluated on full-length videos. The SR scale factor is set to $s = 4$ throughout all experiments. The number of HRBA blocks is $M=4$ for both Net$^D$ and Net$^R$, and the number of multi-flow-mask pairs is $n=9$. For the input to the first HRBA block in Net$^D$, these pairs are initialized with no initial motion and full visibility (\textit{i.e.}, $\bm{f}=\bm{0}$ and $\bm{o}=\bm{1}$). The degradation kernel size is set to $k_d=20$. The loss coefficients (defined in Eqs.~\ref{eq:loss_d} and~\ref{eq:loss_total} of the main paper) are set to $\lambda_1=10^{-4}$, $\lambda_2=10^{-4}$, and $\lambda_3=0.1$. Additionally, we adopt the multi-Dconv head transposed attention (MDTA) and Gated-Dconv feed-forward network (GDFN) modules proposed in Restormer~\cite{zamir2022restormer} for the attention and feed-forward networks in our multi-attention block.

\section{Details of REDS-ME and REDS-RE Benchmarks} \label{sec:supp_datasets}
One significant challenge in VSRDB under dynamic exposures is the lack of benchmarks for systematic performance evaluation. To address this, we construct two new benchmarks, \textbf{REDS-ME} (Multi-Exposure) and \textbf{REDS-RE} (Random-Exposure), derived from the REDS dataset~\cite{nah2019ntire}.

\subsection{Synthesis of Multi-Exposure Sequences (REDS-ME)}

\noindent \textbf{Synthesis Pipeline.} The data generation process approximates the continuous degradation formulation defined in Eq.~\ref{eq:physical_degradation} of the main paper. Following the widely adopted methodology for realistic motion blur synthesis~\cite{nah2017deep, nah2019ntire}, we implement a \textit{blur-then-downsample} pipeline. First, the original 120\,fps REDS videos are interpolated to 1920\,fps using EMA-VFI~\cite{zhang2023extracting} to obtain sufficient intermediate high-framerate frames, denoted as $\bm{H}$.

To simulate the temporal integration of light over a specific exposure time $\Delta t_e$, we average a variable number of these consecutive high-framerate frames. The resulting blurry LR frame $\bm{X}_i$ is formally generated as:
\begin{equation}
    \bm{X}_i \approx \mathcal{D}_s \left( \frac{1}{M_e} \sum_{k=0}^{M_e - 1} \bm{H}[i \cdot M_{int} + k] \right),
    \label{eq:synthesis}
\end{equation}
where $\mathcal{D}_s$ is the spatial bicubic downsampling operator. In the discrete approximation, $M_{int}$ represents the frame interval of the target 24\,fps sequences mapped into the 1920\,fps domain (\ie, $M_{int}=80$), and $M_e$ is the variable number of accumulated high-framerate frames determined by the exposure time $\Delta t_e$. This pipeline isolates exposure duration as the controlled variable governing motion-blur extent. Other camera factors, such as ISO/gain changes, white balance shifts, clipping, and sensor noise, are intentionally outside the scope of REDS-ME/RE; our synthesis follows the standard blur-then-downsample assumption~\cite{nah2017deep, nah2019ntire}.

\noindent \textbf{Definition of Exposure Levels.} To construct REDS-ME, we synthesize five variants of blurry videos representing different exposure levels by varying $M_e$ in Eq.~\ref{eq:synthesis}. Motivated by the original REDS dataset's temporal sampling strategy, we mathematically define the exposure levels based on the duty cycle $r = N/5$, where $N \in \{1, 2, 3, 4, 5\}$. Given the 24\,fps frame interval $\Delta t$, the per-frame exposure time $\Delta t_e$ is precisely defined as:
\begin{equation}
    \Delta t_e = r \Delta t = \frac{N}{5} \Delta t.
\end{equation}
Consequently, the ratio $5\!:\!1$ represents the shortest exposure ($20\%$ duty cycle, minimal motion blur), while $5\!:\!5$ represents the longest exposure ($100\%$ duty cycle, severe motion blur). These precise discrete levels also serve as pseudo-labels for pretraining our ETE module. 

For training, we utilize all five exposure variations from the REDS-ME training set. For evaluation, we adopt the two most challenging exposure levels, $5\!:\!4$ and $5\!:\!5$, from the REDS4-ME test set.

\begin{algorithm}[t]
\caption{Generation of Exposure Trajectory for a Video in REDS-RE}
\label{alg:random_walk}
\begin{algorithmic}[1]
\Require Total frames $T$, Initial exposure level $E_0 \in \{1, 2, 3, 4, 5\}$, Update interval $I \in \{5, 7\}$
\State $E \gets E_0$
\For{$i = 0$ to $T-1$}
    \If{$i == 0$ \textbf{or} $i \bmod I == 0$}
        \State $\Delta E \gets \text{RandomChoice}(\{-1, 0, +1\})$ \Comment{Sample exposure change}
        \State $E \gets \text{Clip}(E + \Delta E, 1, 5)$ \Comment{Update and keep within bounds}
    \EndIf
    \State $\text{Exposure}[i] \gets E$ \Comment{Assign exposure level to current frame}
\EndFor
\State \Return $\text{Exposure}$
\end{algorithmic}
\end{algorithm}

\subsection{Synthesis of Random-Exposure Sequences (REDS-RE)}

To explicitly evaluate robustness under dynamically varying exposure conditions, we construct the REDS-RE benchmark by temporally mixing frames from all five exposure levels within each REDS4-ME test scene. To impose temporal inertia rather than frame-wise randomness, we employ a \textit{structured, interval-based random walk} strategy rather than simple frame-wise randomization.

\begin{figure}[t]
\centering
\includegraphics[width=0.6\linewidth]{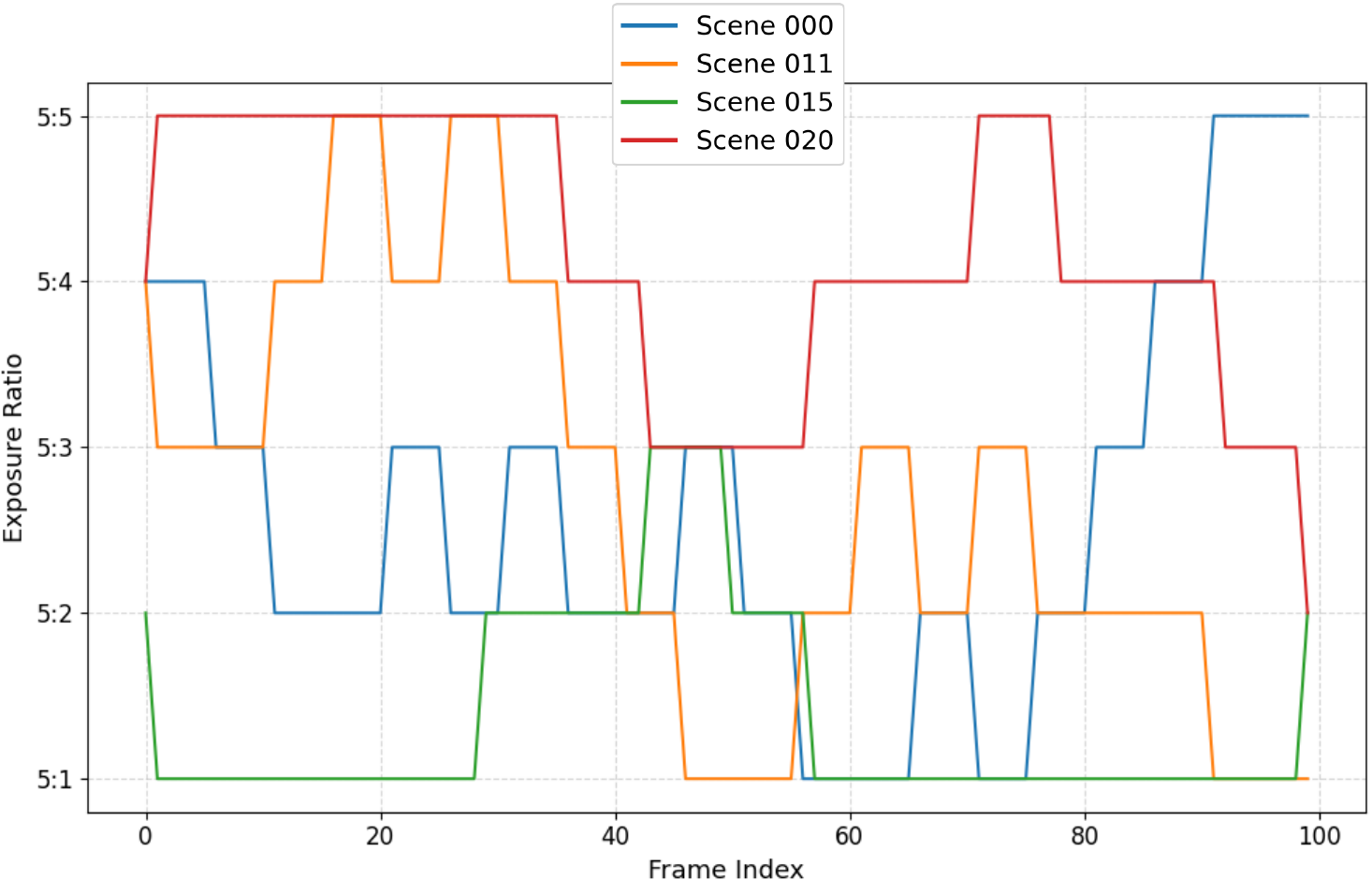}
\caption{Visualization of synthesized exposure trajectories in the REDS-RE benchmark. Each colored line represents the evolution of the exposure level for a different test scene. The trajectories follow a step-wise random walk with varying update intervals, producing temporally coherent exposure-duration variation.}
\label{fig:reds_re_traj}
\end{figure}

Specifically, the exposure level is updated only at fixed intervals. Depending on the test scene, this update interval is set to either every 5 or 7 frames to maintain consistent temporal inertia within a sequence. At each update step, the exposure level is uniformly sampled to increment ($+1$), decrement ($-1$), or remain constant ($0$), constrained within the valid bounds ($5\!:\!1$ to $5\!:\!5$). This generation protocol is detailed in Algorithm~\ref{alg:random_walk}. As visualized in Fig.~\ref{fig:reds_re_traj}, this process yields diverse, step-wise exposure trajectories.

\section{Further Ablation Studies} \label{sec:supp_abla} 

In this section, we provide further ablation studies and detailed analyses that were omitted from the main paper due to space constraints.

\subsection{Quantitative Evaluation of ETE Embeddings} \label{sec:supp_abla_ete}

To demonstrate that our Exposure Time-aware Feature Extractor (ETE) learns a discriminative exposure representation space, we perform a quantitative evaluation of the extracted embeddings $\bm{u}_i$. Note that ETE is optimized purely via a contrastive loss without an explicit classification head. Therefore, we evaluate its inherent separability using a $k$-Nearest Neighbors classifier ($k=5$) with cosine similarity and 5-fold cross-validation on the extracted features.

\begin{figure}[t]
\centering
\includegraphics[width=0.6\linewidth]{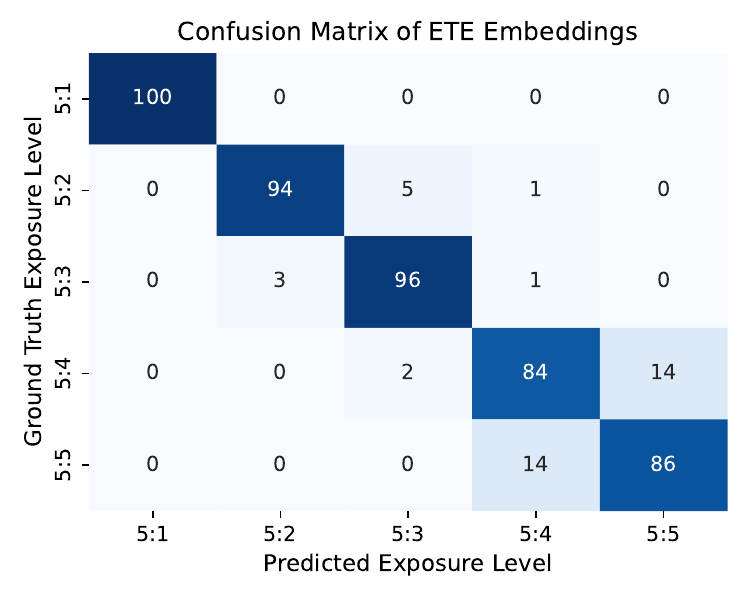}
\caption{Confusion matrix of the learned ETE embeddings, evaluated using a $k$-NN classifier ($k=5$) with cosine similarity. The embeddings achieve a high Top-1 accuracy of 92.0\% with a strongly diagonal distribution. Notably, almost all misclassifications occur between adjacent exposure levels (\textit{e.g.}, between $5\!:\!4$ and $5\!:\!5$), reflecting the inherent visual ambiguity of severe motion blur. Conversely, short exposures (\textit{e.g.}, $5\!:\!1$) are perfectly isolated.}
\label{fig:ete_cm}
\end{figure}

As shown in Fig.~\ref{fig:ete_cm}, the ETE embeddings achieve a high Top-1 classification accuracy of 92.0\%. The confusion matrix reveals a dominantly diagonal structure, confirming that the embeddings are well-separated according to their true exposure levels. Notably, almost all misclassifications are bounded to adjacent exposure levels without any extreme outliers. For instance, the most frequent confusion occurs between the two longest exposures, with 14 samples of $5\!:\!4$ predicted as $5\!:\!5$, and exactly 14 samples of $5\!:\!5$ predicted as $5\!:\!4$. This is expected, as the visual distinction between 80\% and 100\% duty-cycle motion blur is inherently ambiguous (see Fig.~\ref{fig:reds_me} of the main paper). Conversely, short exposures (\textit{e.g.}, $5\!:\!1$), which exhibit minimal blur, are perfectly isolated with 100\% accuracy.

These quantitative results indicate that the ETE successfully captures exposure-dependent structural cues, preserving an ordinal structure consistent with the continuous nature of motion blur. Overall, the ETE provides a reliable exposure-dependent prior for the subsequent restoration networks.

\subsection{Progressive Feature Refinement in HRBA} \label{sec:supp_abla_hrba}

\begin{figure}[h]
\centering
\includegraphics[width=\linewidth]{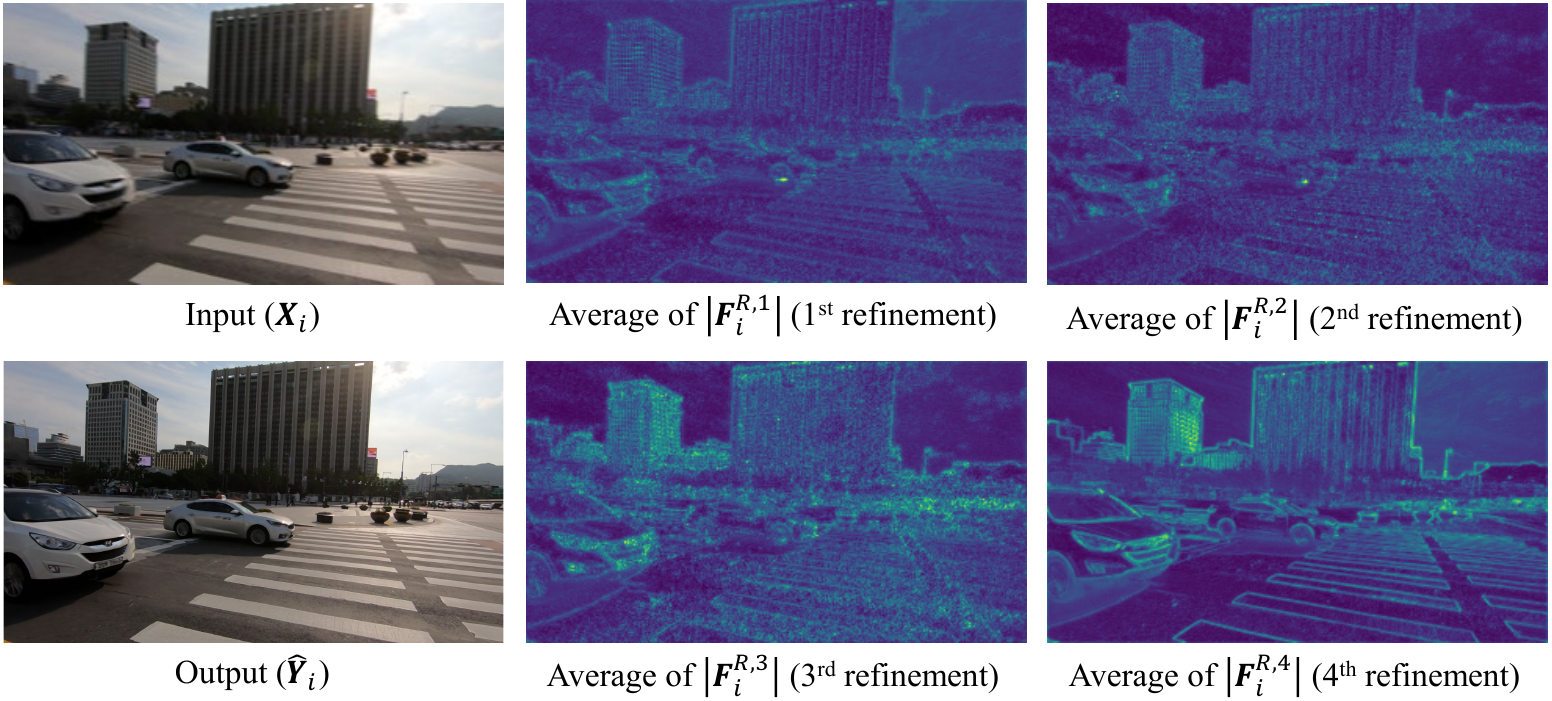}
\caption{Visualization of the progressive feature refinement through HRBA blocks across four iterations.}
\label{fig:hrba_refine_vis}
\end{figure}

We visualize the intermediate representations of the refined feature $\bm{F}_i^{R,j}$ across four refinement stages in Fig.~\ref{fig:hrba_refine_vis} to illustrate how the HRBA blocks progressively operate. As shown in the figure, the initial stage exhibits noisy and spatially diffuse activations, while later stages produce increasingly sharper and more structurally aligned features, with high-frequency details (\textit{e.g.}, building edges) becoming more prominent. This progressive sharpening indicates that our hierarchical refinement strategy iteratively enhances feature quality, leading to sharper and more temporally consistent outputs.

\subsection{Effect of the Number of Multi-Flow-Mask Pairs} \label{sec:supp_abla_n}

We analyze how the number of multi-flow-mask pairs $n$ within our HRBA backbone affects performance and stability in motion estimation. As shown in Table~\ref{tab:supp_abla_n}, increasing $n$ consistently improves restoration quality with negligible computational overhead (only a 0.001s increase in runtime from $n=1$ to $n=9$). A larger $n$ enables the model to establish more robust one-to-many correspondences, effectively leveraging multiple motion hypotheses. This is especially critical under severe motion blur, where a single flow estimation is highly susceptible to localized errors.

\begin{figure}[h]
\centering
\includegraphics[width=\linewidth]{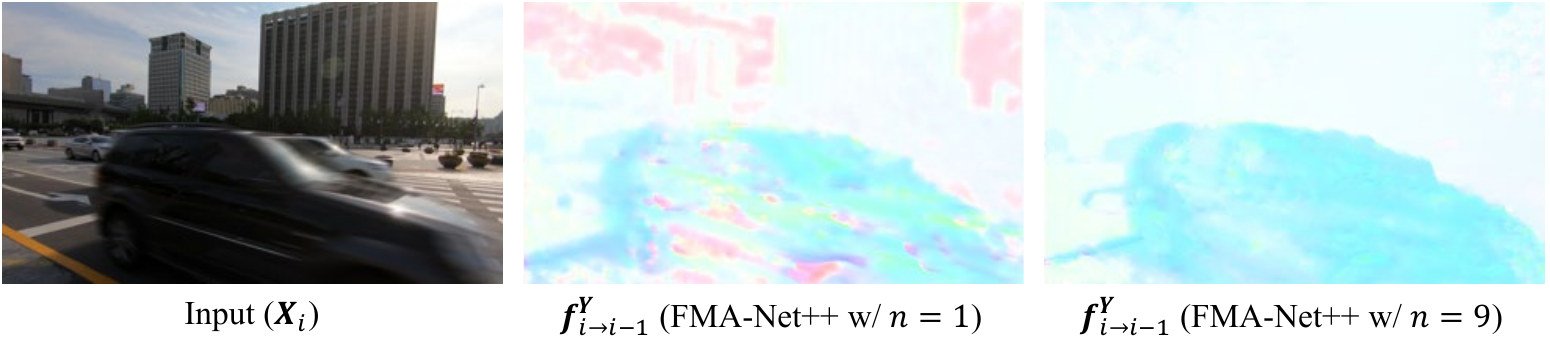}
\caption{Effect of the number of multi-flow-mask pairs ($n$) on the predicted optical flow for a severely blurred scene.}
\label{fig:supp_flow_vis}
\end{figure}

Fig.~\ref{fig:supp_flow_vis} visualizes this effect. With only one pair ($n=1$), the predicted optical flow is noisy and spatially distorted, failing to capture accurate motion boundaries. In contrast, using nine pairs ($n=9$) produces much cleaner and sharper flow fields that align well with actual object motion. This confirms that integrating the multi-flow mechanism remains effective for robust motion modeling under challenging degradation conditions. We thus retain this component and set $n=9$ in our final configuration.

\renewcommand{\arraystretch}{1.1}
\begin{table}[h]
\begin{center}
\caption{Ablation study for the number of multi-flow-mask pairs ($n$) on REDS4-ME-$5\!:\!5$.}
\label{tab:supp_abla_n}
\scalebox{0.88}{
\begin{tabular}{cccc}
\hline
\multicolumn{1}{c|}{\multirow{2}{*}{\# $n$}} & \multicolumn{1}{c|}{\multirow{2}{*}{\# Params (M)}} & \multicolumn{1}{c|}{\multirow{2}{*}{Runtime (s)}} & REDS4-ME-5:5 \\
\multicolumn{1}{c|}{} & \multicolumn{1}{c|}{} & \multicolumn{1}{c|}{} & PSNR$\uparrow$ / SSIM$\uparrow$ / tOF$\downarrow$ \\
\hline
\multicolumn{1}{c|}{$n=1$} & \multicolumn{1}{c|}{11.9} & \multicolumn{1}{c|}{0.073} & 28.52 / 0.8248 / 2.357 \\
\multicolumn{1}{c|}{$n=5$} & \multicolumn{1}{c|}{12.3} & \multicolumn{1}{c|}{0.074} & 28.97 / 0.8387 / 2.106 \\
\multicolumn{1}{c|}{$n=9$} & \multicolumn{1}{c|}{12.8} & \multicolumn{1}{c|}{0.074} & 29.24 / 0.8453 / 1.956 \\
\hline
\end{tabular}}
\end{center}
\vspace{-0.5cm}
\end{table}

\subsection{Effect of Exposure-Aware FGDF} \label{sec:supp_abla_fgdf}

FGDF was originally introduced in FMA-Net~\cite{Youk_2024_CVPR} to perform motion-aware filtering along optical-flow trajectories. In FMA-Net++, we enhance FGDF by explicitly conditioning the filtering weights on the exposure-aware features $\bm{u}$ (Sec.~\ref{sec:arch} of the main paper). To validate the contribution of this exposure conditioning, we compare three variations of degradation modeling on the severely blurred REDS4-ME-$5\!:\!5$ test set: (1) conventional dynamic filtering (CDF)~\cite{jia2016dynamic}, (2) pure FGDF~\cite{Youk_2024_CVPR} (without ETM guidance), and (3) our full exposure-aware FGDF. 

As shown in Table~\ref{tab:supp_fgdf}, while transitioning from CDF to pure FGDF improves degradation modeling by tracking motion trajectories, incorporating the exposure-aware guidance yields a further performance leap across all motion magnitudes. This confirms that our exposure-aware conditioning effectively strengthens the underlying motion-aware modeling, providing accurate degradation kernels even in challenging high-motion scenarios.

\renewcommand{\arraystretch}{1.1}
\begin{table}[h]
\begin{center}
\caption{Comparison of different degradation modeling mechanisms on REDS4-ME-$5\!:\!5$, reporting the Net$^D$ prediction performance. Each cell reports PSNR\slash tOF values averaged within each motion magnitude interval.}
\label{tab:supp_fgdf}
\scalebox{0.78}{
\begin{tabular}{lccc}
\toprule
Network: Net$^D$ & [0,20) & [20,40) & $\ge40$ \\
\midrule
CDF~\cite{jia2016dynamic} & 47.67 / 0.046 & 42.92 / 0.228 & 34.99 / 0.688 \\
FGDF~\cite{Youk_2024_CVPR} (w/o ETM) & 48.42 / 0.042 & 43.89 / 0.206 & 36.87 / 0.651 \\
Exposure-aware FGDF (Ours) & \textbf{48.57} / \textbf{0.040} & \textbf{44.21} / \textbf{0.197} & \textbf{37.38} / \textbf{0.637} \\
\bottomrule
\end{tabular}}
\end{center}
\vspace{-0.5cm}
\end{table}

\subsection{Analysis of Loss Functions} \label{sec:supp_abla_loss}

We validate the design of our composite loss function $\mathcal{L}_D$ (Eq.~\ref{eq:loss_d} of the main paper), which guides the training of Net$^D$. Specifically, we analyze the impact of the coefficients for the warping loss ($\lambda_1$) and the RAFT supervision loss ($\lambda_2$) on the REDS4-ME-$5\!:\!5$ test set. 

As summarized in Table~\ref{tab:abla_loss}, both loss terms are essential for achieving optimal performance. First, adjusting the weight ($\lambda_1$) of the warping loss term significantly affects the final restoration quality: an overly large weight interferes with the primary reconstruction objective, while a weight that is too small fails to enforce accurate alignment in the sharp HR space. Second, removing the RAFT supervision ($\lambda_2 = 0$) causes a notable drop in performance, confirming that pseudo-GT flow supervision is crucial for learning accurate motion priors. Our chosen coefficients ($\lambda_1=\lambda_2=10^{-4}$) provide the best trade-off, yielding the highest performance across all metrics.

\renewcommand{\arraystretch}{1.1}
\begin{table}[h]
\begin{center}
\caption{Ablation on the loss coefficients ($\lambda_1$ and $\lambda_2$) used in $\mathcal{L}_D$ evaluated on the REDS4-ME-$5\!:\!5$ test set.}
\label{tab:abla_loss}
\scalebox{0.85}{
\begin{tabular}{lcc}
\toprule
Hyperparameters & \multicolumn{2}{c}{REDS4-ME-5:5} \\
 & PSNR $\uparrow$ / SSIM $\uparrow$ & tOF $\downarrow$ \\
\midrule
\multicolumn{3}{c}{\textit{Analysis on Warping Loss ($\lambda_1$)}} \\
\midrule
$\lambda_1 = 10^{-3}$      & 29.13 / 0.8395 & 2.013 \\
$\lambda_1 = 5 \times 10^{-5}$ & 29.20 / 0.8437 & 1.971 \\
\midrule
\multicolumn{3}{c}{\textit{Analysis on RAFT Supervision ($\lambda_2$)}} \\
\midrule
$\lambda_2 = 0$ (w/o RAFT) & 29.07 / 0.8347 & 2.143 \\
$\lambda_2 = 10^{-3}$      & 29.12 / 0.8391 & 2.022 \\
$\lambda_2 = 5 \times 10^{-5}$ & 29.16 / 0.8409 & 1.998 \\
\midrule
$\lambda_1=10^{-4}, \lambda_2=10^{-4}$ (Final) & \textbf{29.24} / \textbf{0.8453} & \textbf{1.956} \\
\bottomrule
\end{tabular}}
\end{center}
\vspace{-0.5cm}
\end{table}

\section{Additional Qualitative Results} \label{sec:supp_qual}
In this section, we provide extended visual comparisons and analyses to complement the quantitative results presented in the main paper.

\subsection{Qualitative Comparison with FMA-Net} \label{sec:supp_qual_fmanet}

While Tables~\ref{tab:reds_me4_comparison} and~\ref{tab:gopro_comparison} of the main paper demonstrate that our FMA-Net++ quantitatively outperforms the retrained FMA-Net$^*$~\cite{Youk_2024_CVPR} (which utilizes a sliding-window approach), we further provide a direct visual comparison to highlight their structural restoration capabilities. We specifically focus on challenging scenes containing severe motion blur and low spatial redundancy (\textit{e.g.}, human faces), where leveraging rich, long-range temporal context is critical to compensate for the heavily degraded spatial information.

\begin{figure}[h]
\centering
\includegraphics[width=\linewidth]{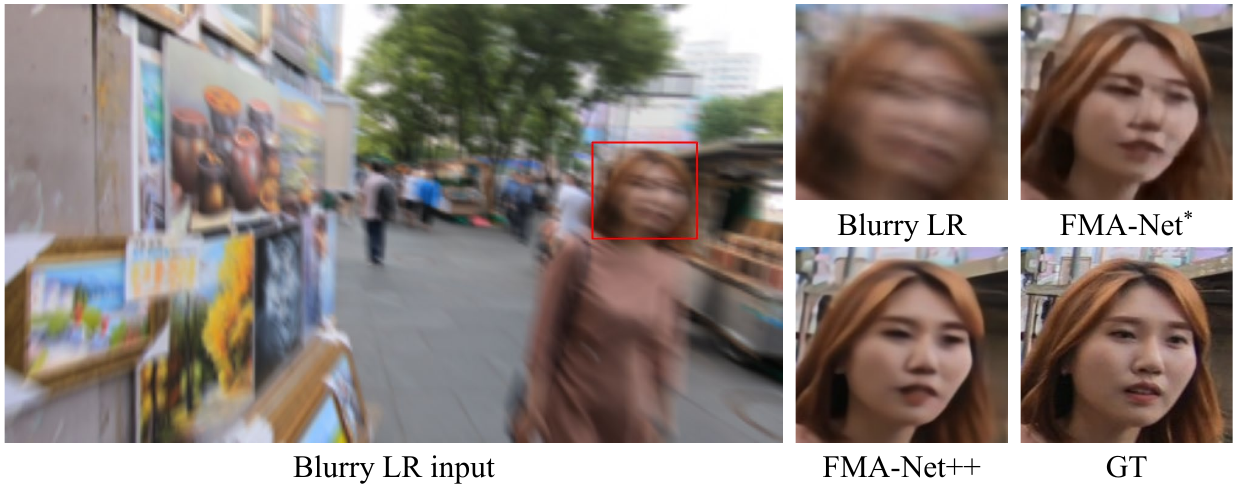}
\caption{Qualitative comparison between FMA-Net$^*$~\cite{Youk_2024_CVPR} and FMA-Net++ (Ours) in a challenging scene featuring complex facial details and severe motion blur. Our model successfully reconstructs sharp and temporally consistent structures without distortion.}
\label{fig:fmanet_comparison}
\vspace{-0.5cm}
\end{figure}

As shown in Fig.~\ref{fig:fmanet_comparison}, FMA-Net$^*$, constrained by its narrow temporal window ($T=3$), fails to gather sufficient temporal redundancy from adjacent frames and consequently produces over-smoothed textures and distorted facial structures. In contrast, our FMA-Net++ utilizes the HRBA backbone to hierarchically aggregate and refine features, effectively leveraging a broader temporal context. This enhanced temporal modeling allows our model to restore sharper edges and more temporally consistent, high-frequency details. These visual results show that our hierarchical design overcomes the temporal limitations of the baseline FMA-Net framework.

\subsection{Additional Visual Comparisons and Real-World Generalization} \label{sec:supp_qual_additional}

\begin{figure*}[t]
\centering
\includegraphics[width=\linewidth]{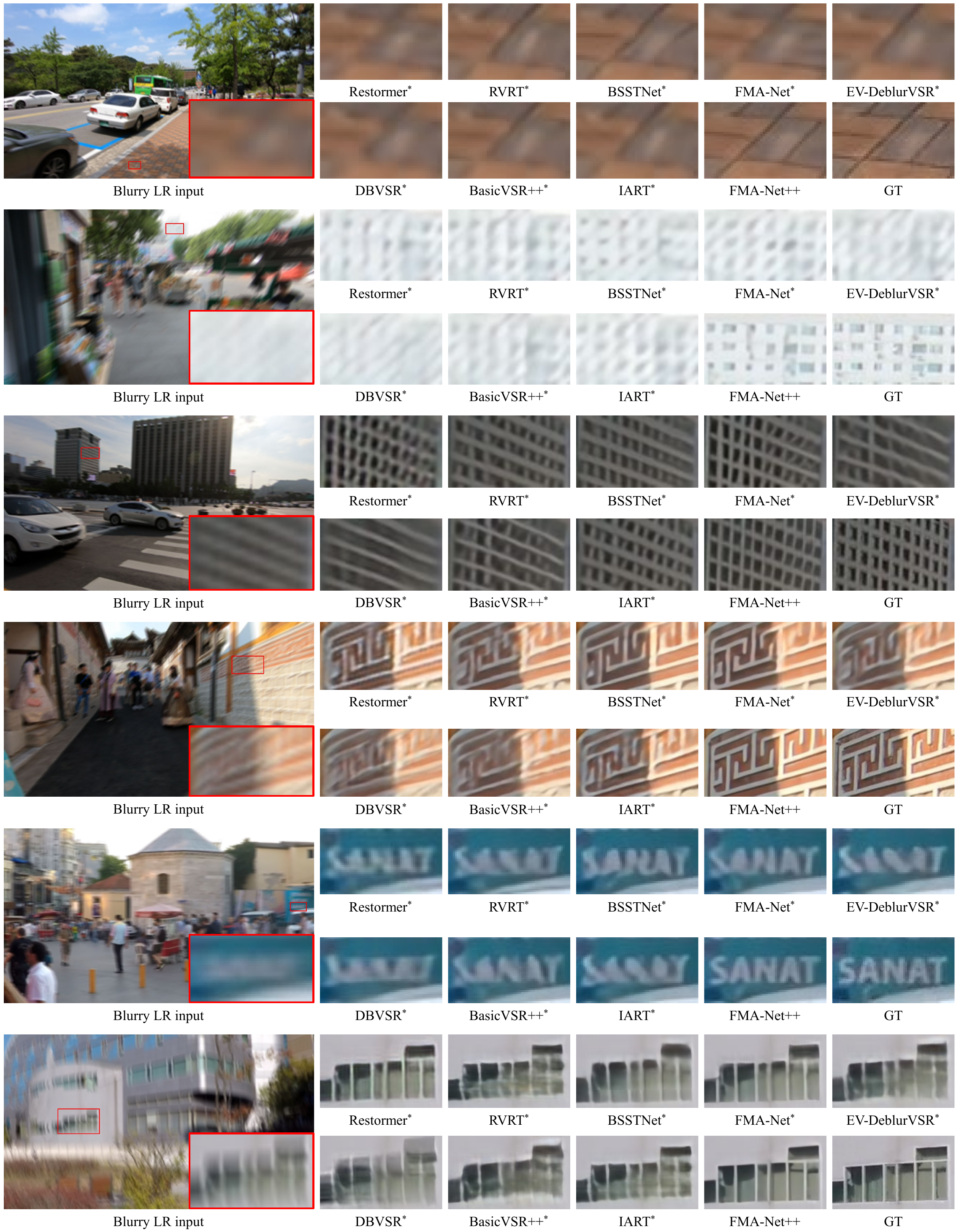}
\caption{Additional qualitative comparisons on the REDS4-ME-$5\!:\!5$ and GoPro~\cite{nah2017deep} test sets. These scenes feature severe motion blur and complex textures, representing highly challenging degradation scenarios. Compared to existing state-of-the-art methods, FMA-Net++ consistently reconstructs sharper structural details and cleaner edges while effectively suppressing severe motion artifacts. \textit{Best viewed in zoom}.}
\label{fig:supp_reds} 
\end{figure*}

We provide additional qualitative comparisons to complement the results shown in the main paper. Further visual results on challenging scenes from the REDS4-ME-$5\!:\!5$ and GoPro test sets are shown in Fig.~\ref{fig:supp_reds}. 

\begin{figure*}[t]
\centering
\includegraphics[width=\linewidth]{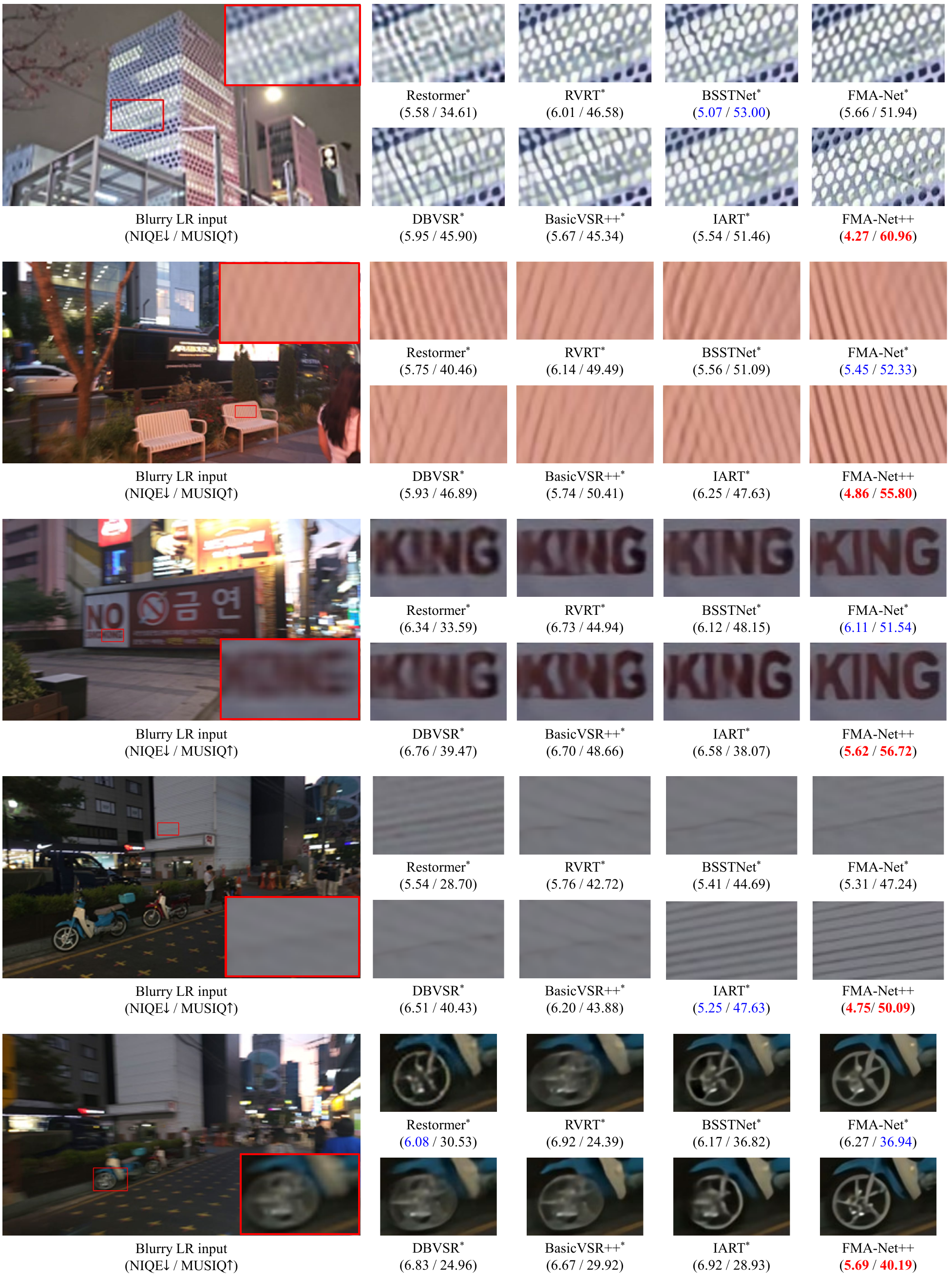}
\caption{Qualitative comparisons on challenging real-world videos captured with smartphones. These videos contain continuous exposure changes and non-uniform motion blur, deviating from the discrete synthetic anchors used during training. Despite this domain gap, FMA-Net++ recovers legible text and fine textures, and achieves favorable no-reference scores (NIQE$\downarrow$ / MUSIQ$\uparrow$) in these examples. \textit{Best viewed in zoom}.}
\label{fig:supp_real_world}
\end{figure*}

Furthermore, examples of real-world video restoration are presented in Fig.~\ref{fig:supp_real_world}. These real-world smartphone videos include continuous exposure changes and non-uniform motion blur that are not restricted to the discrete synthetic anchors ($5\!:\!1$ to $5\!:\!5$) used during training. Although the ETE is optimized only on discrete exposure anchors, FMA-Net++ still restores these real-world videos well. The ordinal structure analyzed in Sec.~\ref{sec:supp_abla_ete} suggests that the learned exposure-aware feature space can interpolate between nearby exposure conditions, helping the model handle intermediate, unseen exposure states.

\section{Limitations and Future Work} \label{sec:supp_limit}
While FMA-Net++ addresses the coupled degradation of motion blur and dynamic exposure, our current formulation and REDS-ME/RE benchmarks deliberately focus on blur-extent variation induced by temporal integration. They do not explicitly model other entangled camera factors, such as ISO/gain changes, white-balance shifts, clipping, dynamic-range changes, heavy sensor noise, or spatially varying lighting. Extending our framework to jointly handle these complex, concurrent factors presents an important direction for future research.

\end{sloppypar}

\end{document}